\documentclass[10pt,twocolumn,letterpaper]{article}

\usepackage{iccv}
\usepackage{times}
\usepackage{epsfig}
\usepackage{graphicx}
\usepackage{amsmath}
\usepackage{amssymb}

\usepackage{mathtools}

\usepackage{times}
\usepackage{epsfig}
\usepackage{graphicx}
\usepackage{amsmath}
\usepackage{amssymb}
\usepackage{verbatim}
\usepackage{hhline}
\usepackage{multirow}
\usepackage{makecell}

\usepackage{amssymb}
\usepackage{pifont}
\newcommand{\cmark}{\ding{51}}%

\newcommand{\Tref}[1]{Table~\textcolor{blue}{\ref{#1}}}

\newcommand{\Fref}[1]{Fig.~\textcolor{blue}{\ref{#1}}}
\newcommand{\Sref}[1]{Sec.~\textcolor{blue}{\ref{#1}}}
\newcommand{\Aref}[1]{Alg.~\textcolor{blue}{\ref{#1}}}

\usepackage[dvipsnames]{xcolor}

\usepackage{algorithm}
\usepackage{algpseudocode}

\usepackage[pagebackref=true,breaklinks=true,letterpaper=true,colorlinks,bookmarks=false]{hyperref}

\iccvfinalcopy 

\def\iccvPaperID{9550} 

\ificcvfinal\fi

\begin{document}

\title{VolumeFusion: Deep Depth Fusion for 3D Scene Reconstruction}

\author{
Jaesung Choe\\
KAIST\\
\and
Sunghoon Im\\
DGIST\\
\and
Francois Rameau\\
KAIST\\
\and
Minjun Kang\\
KAIST\\
\and
In So Kweon\\
KAIST\\
}

\maketitle

\begin{abstract}
To reconstruct a 3D scene from a set of calibrated views, traditional multi-view stereo techniques rely on two distinct stages: local depth maps computation and global depth maps fusion. Recent studies concentrate on deep neural architectures for depth estimation by using conventional depth fusion method or direct 3D reconstruction network by regressing Truncated Signed Distance Function (TSDF). In this paper, we advocate that replicating the traditional two stages framework with deep neural networks improves both the interpretability and the accuracy of the results. As mentioned, our network operates in two steps: 1) the local computation of the local depth maps with a deep MVS technique, and, 2) the depth maps and images' features fusion to build a single TSDF volume. In order to improve the matching performance between images acquired from very different viewpoints (\eg, large-baseline and rotations), we introduce a rotation-invariant 3D convolution kernel called PosedConv. The effectiveness of the proposed architecture is underlined via a large series of experiments conducted on the ScanNet dataset where our approach compares favorably against both traditional and deep learning techniques.

\end{abstract}

\section{Introduction}
\label{sec:Introduction}

Multi-view stereo (MVS) is a fundamental research topic that has been extensively investigated over the past decades~\cite{geometry}. The main goal of MVS is to reconstruct a 3D scene from a set of images acquired from different viewpoints. This problem is commonly framed as a correspondence search problem by optimizing photometric or geometric consistency among groups of pixels in different images.
Conventional non-learning-based MVS frameworks~\cite{gallup2007real,campbell2008using,furukawa2009accurate} generally achieve the reconstruction using various 3D representations~\cite{intro_mvs}: depth maps~\cite{gallup2007real,campbell2008using}, point cloud~\cite{furukawa2009accurate}, voxels~\cite{lorensen1987marching,hernandez2007probabilistic}, and meshes~\cite{esteban2004silhouette}.

The recent use of deep neural networks for MVS~\cite{mvdepthnet,gpmvs,dpsnet,atlas,mvsnet,point_mvs,cascade_mvs} has proven effective in addressing the limitations of traditional techniques like repetitive patterns, low-texture regions, and reflections. Usually, deep-based MVS methods~\cite{mvsnet,point_mvs,cascade_mvs} center on the estimation of the pixel-wise correspondence between a reference image and its surrounding views. While this strategy can be elegantly integrated into deep learning frameworks, they can only work locally where frames largely overlap. For the full 3D reconstruction of the entire scene, these methods~\cite{mvsnet,point_mvs,cascade_mvs} require to perform a depth map fusion~\cite{openmvg,galliani2015massively} to merge local reconstructions as post-processing.

\begin{figure}[!t]
\centering
\includegraphics[width=0.99\linewidth]{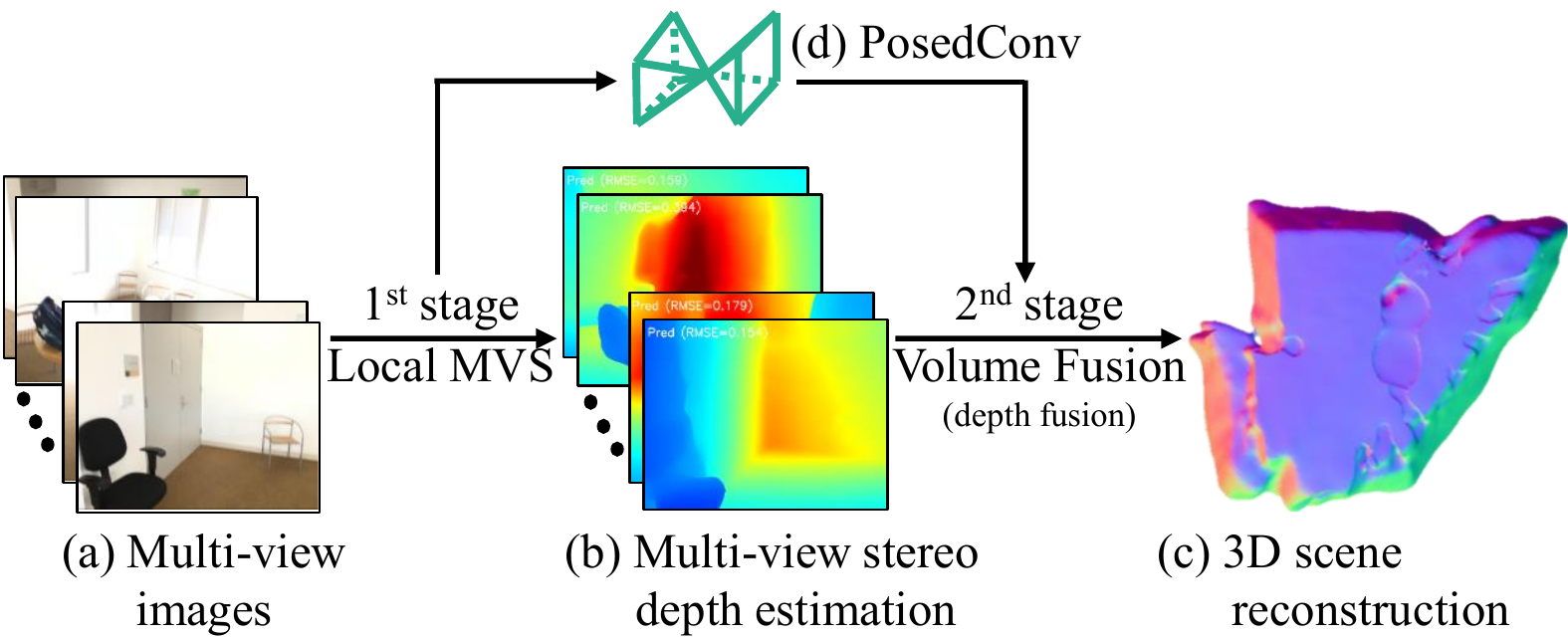}
\caption{\textbf{Volume fusion.} 
Given (a) multi-view images and their camera parameters,
our network aims at 3D scene reconstruction. (b) First, we estimate local multi-view depth maps. (b) Second, we introduce differentiable depth fusion with the guidance of pose-invariant features from (d) our PosedConv.} 
\label{fig:fig_main}
\end{figure}

More recently, Murez~\etal~\cite{atlas} suggest that a direct regression of the Truncated Signed Distance Function (TSDF) volume of the scene is more effective than using intermediate 3D representations,~\ie depth maps. The overall concept of their technique consists in the back-projection of all the extracted image features into a global scene volume from which the network directly regresses the TSDF volume. This end-to-end approach~\cite{atlas} has the advantage of being straightforward and scalable to a large number of images. However, this pioneering work~\cite{atlas} has difficulty in shaping the global structure of complex scenes, such as the corners of rooms or long hallways.

To address this issue, we propose to closely mimic the traditional 3D reconstruction pipeline with two distinct stages: local reconstruction and global fusion. 
However, unlike the previous studies~\cite{mvsnet,deepvideomvs,atlas} and concurrent papers~\cite{neural_recon,trans_fusion}, we integrate these two stages in an end-to-end manner. First, our network computes the local geometry,~\ie, dense depth maps from neighboring frames. Then, we begin the depth fusion process by merging local depth maps as well as image features in a single volume representation where our network regresses TSDF for the final 3D scene reconstruction. This enables our end-to-end framework to learn a globally consistent volumetric representation in a single forward computation without the need for manually engineered fusion algorithms~\cite{colmap,galliani2015massively}. To further enhance the robustness of our depth fusion mechanism, we propose the \emph{Posed Convolution Layer} (Posed-Conv). Compared to the traditional 3D convolution layer that is solely invariant to translation, we propose a more versatile convolution layer that is invariant to both translation and rotation. In short, our Posed Convolution Layer helps to extract pose-invariant feature representation regardless of the orientation of the input image. As a result, our method demonstrates globally consistent shape reconstruction even under wide baselines or large rotations between the views. Our contributions are summarized as follows:
\begin{itemize}
\item A novel network taking advantage of local~MVS and global~depth fusion for 3D scene reconstruction.
\item A new rotation and translation invariant convolution layer, called \emph{PosedConv}.
\end{itemize}

\section{Related Works}
\label{sec:Related works}

\subsection{Multi-view Stereo}
Multi-View Stereo (MVS) consists in the pixel-wise 3D reconstruction of a scene given a set of unstructured images along with their respective intrinsic and extrinsic parameters. In~\cite{intro_mvs}, the scene representation is used as an axis of taxonomy to categorize MVS into four sub-fields of research: depth maps~\cite{gallup2007real,campbell2008using}, point clouds~\cite{furukawa2009accurate,old_mvs_point_cloud_00}, voxels~\cite{lorensen1987marching,hernandez2007probabilistic}, and meshes~\cite{esteban2004silhouette}. In particular, The depth map estimation approaches~\cite{gallup2007real,campbell2008using,plane_sweep_02,plane_sweep_03} have been widely researched since these strategies are easily scalable to number of multi-view images. These methods usually rely on a small baseline assumption to ensure large overlap with a reference frame. Thus, the photometric matching is achieved via a plane-sweeping algorithm~\cite{plane_sweep_00,plane_sweep_01,plane_sweep_02,plane_sweep_03} where the most probable depth is estimated for each pixel in the reference frame. Then, a depth map fusion algorithm~\cite{hernandez2007probabilistic,galliani2015massively} is required to build a global 3D model from a set of depths.

With the rise of deep neural networks, learning-based MVS methods have achieved promising results. Inspired by stereo matching networks~\cite{gcnet,psmnet}, MVS studies~\cite{mvsnet,point_mvs,cascade_mvs,dpsnet,gpmvs,mvdepthnet} have developed cost volume for unstructured multi-view matching. Relying on basic frameworks, such as DPSNet~\cite{dpsnet} or MVSNet~\cite{mvsnet}, follow-up research proposes point-based depth refinement~\cite{point_mvs}, cascaded depth refinement~\cite{cascade_mvs}, and temporal fusion network~\cite{gpmvs,deepvideomvs}. After exhaustively estimating a collection of depth maps, depth fusion~\cite{galliani2015massively,curless1996volumetric} starts to reconstruct the global 3D scene.

\subsection{Depth Map Fusion}
In their seminal work, Curless and Levoy~\cite{curless1996volumetric} propose a volumetric depth map fusion approach able to deal with noisy depth maps through cumulative weighted signed distance function. Follow-up research, such as KinectFusion~\cite{izadi2011kinectfusion} or voxel hashing~\cite{niessner2013real,kahler2015hierarchical}, concentrate on the problem of volumetric representation via depth maps fusion. Recently, with the help of deep learning networks, learning-based volumetric approaches~\cite{ji2017surfacenet,paschalidou2018raynet,routedfusion} have been proposed. For instance, SurfaceNet~\cite{ji2017surfacenet} and RayNet~\cite{paschalidou2018raynet} infer the depth maps from multi-view images and their known camera poses. These methods are close to our strategy, but their networks are trained only by a per-view depth map which is not directly related to depth maps fusion. More recently, RoutedFusion~\cite{routedfusion} and Neural Fusion~\cite{neuralfusion} introduce a new learning-based depth map fusion using RGB-D sensors. However, these papers~\cite{routedfusion,neuralfusion} concentrate on depth fusion algorithm using noisy and uncertain depth maps from multi-view images, not RGB-D sensors. To the best of our knowledge, our method is the first learning-based depth maps fusion from multi-view images for 3D reconstruction.

\begin{figure*}[!t]
\centering
\includegraphics[width=1.00\linewidth]{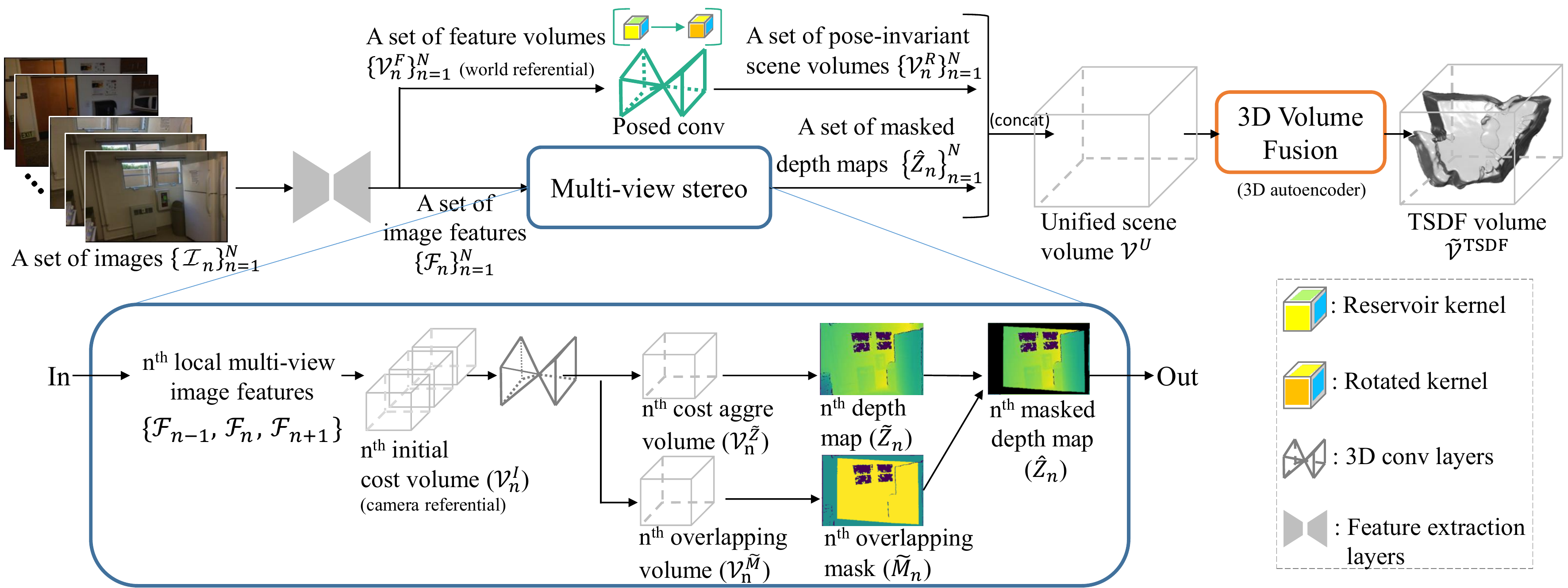}
\caption{\textbf{Overall architecture.} Our volume fusion network consists in two stages: multi-view stereo and Volumetric Depth Fusion. Given a set of $N$ images $\{\mathcal{I}_{n}\}_{n\text{=}1}^{N}$, we extract image features $\{\mathcal{F}_{n}\}_{n\text{=}1}^{N}$. These image features are used (1) to infer a masked depth map $\hat{\mathcal{Z}}_{n}$ in the multi-view stereo stage, and (2) to extract pose-invariant feature volume $\{\mathcal{V}_{n}^{R}\}_{n\text{=}1}^{N}$ in~\emph{PosedConv}. Then, we construct a unified scene volume $\mathcal{V}^{U}$ from $\{\mathcal{V}_{n}^{R}\}_{n\text{=}1}^{N}$ and $\{\hat{\mathcal{Z}}_{n}\}_{n\text{=}1}^{N}$. Finally, in volume fusion stage, we obtain a TSDF volume $\tilde{\mathcal{V}}^\text{TSDF}$ which is the full 3D scene reconstruction results.}
\label{fig:fig_arch}
\end{figure*}

\section{Volume Fusion Network}
\label{sec:Volume Fusion Network}
In this work, we present a novel strategy to effectively reconstruct 3D scenes from an arbitrary set of images~$\mathcal{I}$ where the cameras' parameters for each frame are assumed to be known. The cameras' parameters include the intrinsic matrix $\mathbf{K}$ and the extrinsic parameters (rotation matrix $\mathbf{R} {\in} \mathbb{R}^{3 \times 3}$ and translation vector $\mathbf{t} {\in} \mathbb{R}^{3 \times 1}$). To achieve this reconstruction, we design a novel volume fusion network that consists of two stages: First, each image in the set is processed through our depth network~(\Sref{subsec:Multi-View Stereo}). Based on a pixel-wise photometric matching between adjacent frames, this network regresses both a depth map~$\tilde{\mathcal{Z}}$ and an overlapping mask~$\tilde{\mathcal{M}}$ for every frame. Second, using the resulting per-frame depth maps and overlapping masks, we formulate a depth fusion process as a volume fusion~(\Sref{subsec:Volumetric Depth Fusion}). We further enhance the features extracted from each posed image through our PosedConv~(\Sref{subsec:Posed Convolution Layer}). The overall architecture is presented as in~\Fref{fig:fig_arch}.

\subsection{Multi-view Stereo}
\label{subsec:Multi-View Stereo}
In this section, we describe the first stage of our approach: the local MVS network using three neighbor frames~$\{ \mathcal{I}_{\text{n}\text{-}1}, \mathcal{I}_\text{n}, \mathcal{I}_{\text{n}\text{+}1} \}$. Following previous studies~\cite{dpsnet,normal_assited}, we construct an initial cost volume~$\mathcal{V}^{\mathcal{I}}_\text{n}$ using the neighbor image feature maps~$\{ \mathcal{F}_{\text{n}\text{-}1}, \mathcal{F}_\text{n}, \mathcal{F}_{\text{n}\text{+}1} \}$ to infer a depth map in the reference camera view ($\mathcal{I}_\text{n}$). In contrast to the previous studies~\cite{mvsnet,point_mvs,cascade_mvs} that only concentrate on accurate depth estimation, our network additionally infers an overlapping mask. The overlapping mask~$\tilde{\mathcal{M}}_\text{n}$\footnote{We visualize an overlapping mask in the supplementary material.} is computed to quantify the probability of per-pixel overlap among three adjacent frames in a referential camera view. The purpose of the overlapping mask is to filter out the uncertain depth values that we cannot geometrically deduce,~\ie, without correspondence across neighbor frames.

In details, we compute matching cost by computing the initial cost volume~$\mathcal{V}_\text{n}^{\mathcal{I}}$ through stacked hourglass networks~\cite{associate_embedding,hourglass,psmnet}. Then, we obtain a cost aggregated volume\footnote{We represent a 3D dimensional volume with its channels as four axes: channel~($\text{C}$), depth~($\text{D}$), height~($\text{H}$), and width~($W$).}~$\mathcal{V}^{\tilde{\mathcal{Z}}}_\text{n} {\in} \mathbb{R}^{1 {\times} \text{D} {\times} \text{H} {\times} \text{W}}$ and an overlapping volume~$\mathcal{V}_\text{n}^{\tilde{\mathcal{M}}} {\in} \mathbb{R}^{2 {\times} \text{D} {\times} \text{H} {\times} \text{W}}$. These volumes are used for the inference of a depth map~$\tilde{\mathcal{Z}} {\in} \mathbb{R}^{\text{H} {\times} \text{W}}$ and an overlapping mask~$\tilde{\mathcal{M}} {\in} \mathbb{R}^{\text{H} {\times} \text{W}}$. Regarding the estimation of the overlapping mask, we formulate its estimation as a binary classification problem (\ie, overlap/non-overlap). First, the overlapping probability $\mathcal{P}^{\tilde{\mathcal{M}}}_\text{n}$ ${\in} \mathbb{R}^{1 {\times} \text{D} {\times} \text{H} {\times} \text{W}}$ of the overlapping volume~$\mathcal{V}_\text{n}^{\tilde{\mathcal{M}}}$ is obtained by a softmax operation. Then, the overlapping mask $\Tilde{M}_\text{n}$ can be directly estimated by a max-pooling operation along the depth axis of the probability volume. Specifically, the overlapping probability $\tilde{\mathcal{M}}_{u,v}$ at the pixel location $(u,v)$ can be estimated as follow:
\begin{equation}
\begin{gathered}
    \tilde{\mathcal{M}}_{u,v} {=} \text{MaxPool}_{d \in D} \ (\mathcal{P}^{\tilde{\mathcal{M}}}_\text{n}[d,v,u]),
\label{eq:get-conf-map}
\end{gathered}
\end{equation}
where $\mathcal{P}^{\tilde{\mathcal{M}}}_\text{n}[d,v,u]$ is the probability of the overlapping volume at the voxel $[d,v,u]$. To learn the overlapping mask, a per-pixel L1-Loss between the ground-truth and the estimated mask is employed: 
\begin{equation}
\begin{gathered}
    \mathcal{L}_{M} = \sum_{(u,v) {\in} \tilde{\mathcal{M}}} \| \tilde{\mathcal{M}}_{u,v} - {\mathcal{M}}_{u,v} \|_{1},
\label{eq:conf-loss}
\end{gathered}
\end{equation}
where $\mathcal{L}_{M}$ is an overlapping loss and ${\mathcal{M}}_{u,v}$ is a true overlapping mask $\mathcal{M}$ at pixel $(u,v)$. Note that the overlapping mask is an essential element for our depth map fusion stage (see~\Sref{subsec:Volumetric Depth Fusion}).

Apart from the overlapping mask, the local depth map $\Tilde{z}_\text{n}$ is also compute from the cost aggregated volume $\mathcal{V}^{\tilde{\mathcal{Z}}}_\text{n}$. The depth estimation $\Tilde{z}_{u, v}$ at pixel $(u, v)$ is computed as follows: 
\begin{equation}
    \Tilde{z}_{u, v} = \sum_{d\text{=}1}^\text{D} \frac{z_\text{min} \times D}{d} \cdot \sigma ({\mathbf{a}}_{u,v}^{d}),
\end{equation}
where $z_\text{min}$ is a hyper-parameter defining the minimum range of depth estimation, $\sigma(\cdot)$ represents the softmax operation, ${\mathbf{a}}_{u,v}^{d}$ is the $d$-th plane value of the vector at the pixel $(u, v)$ in the cost aggregated  volume~$\mathcal{V}^{\tilde{\mathcal{Z}}}_\text{n}$. We set the hyper-parameters as $D\text{=}48$ and $z_\text{min}\text{=}0.5~\text{meter}$ in our experiments. We compute the depth loss~$\mathcal{L}_{\mathcal{Z}}$ from the estimated depth map~$\tilde{\mathcal{Z}}$ and a true depth map~$\mathcal{Z}$ as follows: 
\begin{equation}
    \mathcal{L}_{\mathcal{Z}} = \sum_{u}\sum_{v} \mathcal{M}_{u,v} \times \text{smooth}_{\text{L}_{1}}({z}_{u,v} - \tilde{z}_{u,v}),
    \label{equation:disp-loss}
\end{equation}
where $\tilde{z}_{u,v}$ is the value of the predicted depth map $\widetilde{\mathcal{Z}}$ at pixel~$(u, v)$, and~$\text{smooth}_{\text{L}_{1}}(\cdot)$ is the smooth L1-loss function. Note that we mask out the depth map with the ground-truth mask $\mathcal{M}$ to impose the loss only on the pixels where the correspondence among the neighbor views exists.

\begin{figure*}[!t]
\centering
\includegraphics[width=1.00\linewidth]{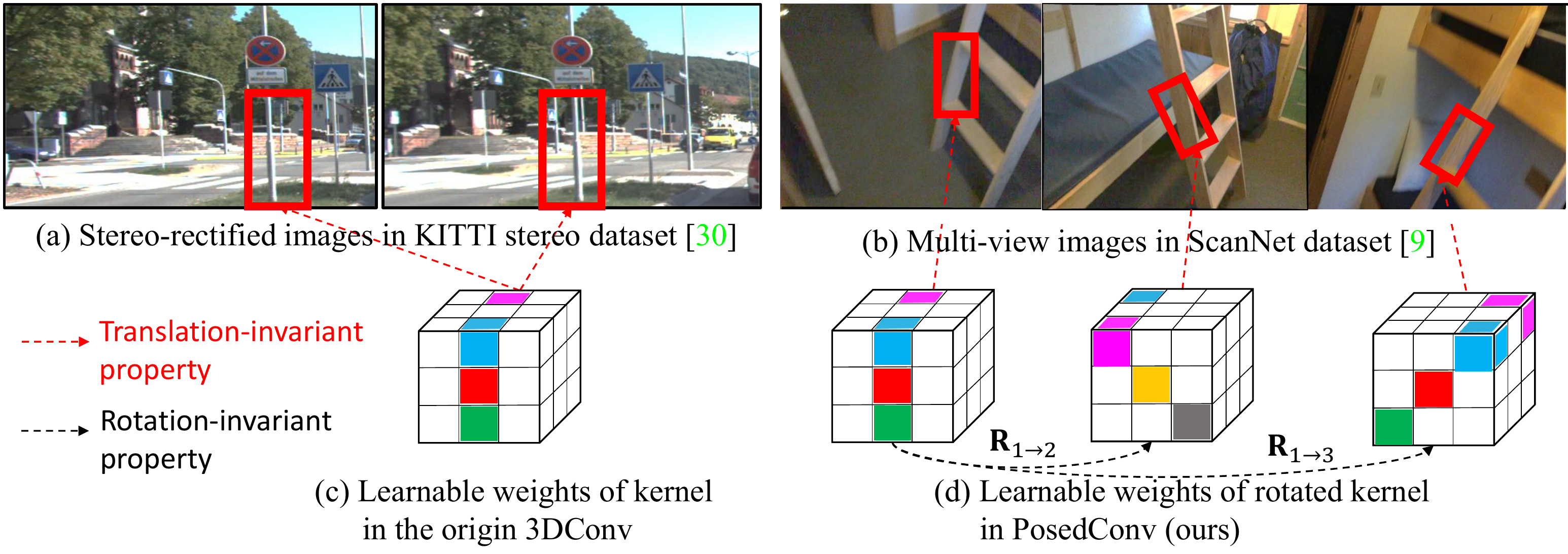}
\caption{\textbf{Illustration of origin 3DConv and our PosedConv layer.} Compared to (a)~the stereo-rectified images~\cite{kitti-stereo}, (b) multi-view images~\cite{scannet} describe the identical objects (red boxes) in different viewpoints. Accordingly, (c)~the original 3DConv can be used in cost volumes~\cite{psmnet,ganet} to compute the matching cost among the stereo-rectified images. (d) Our PosedConv has both rotation-invariant and translation-invariant property that properly extracts features from differently posed images for the robust matching in the world-referential coordinate (\ie, unified scene volume $\mathcal{V}^{U}$).}
\label{fig:fig_posed_conv}
\end{figure*}

\begin{figure}[!t]
\centering
\includegraphics[width=1.0\linewidth]{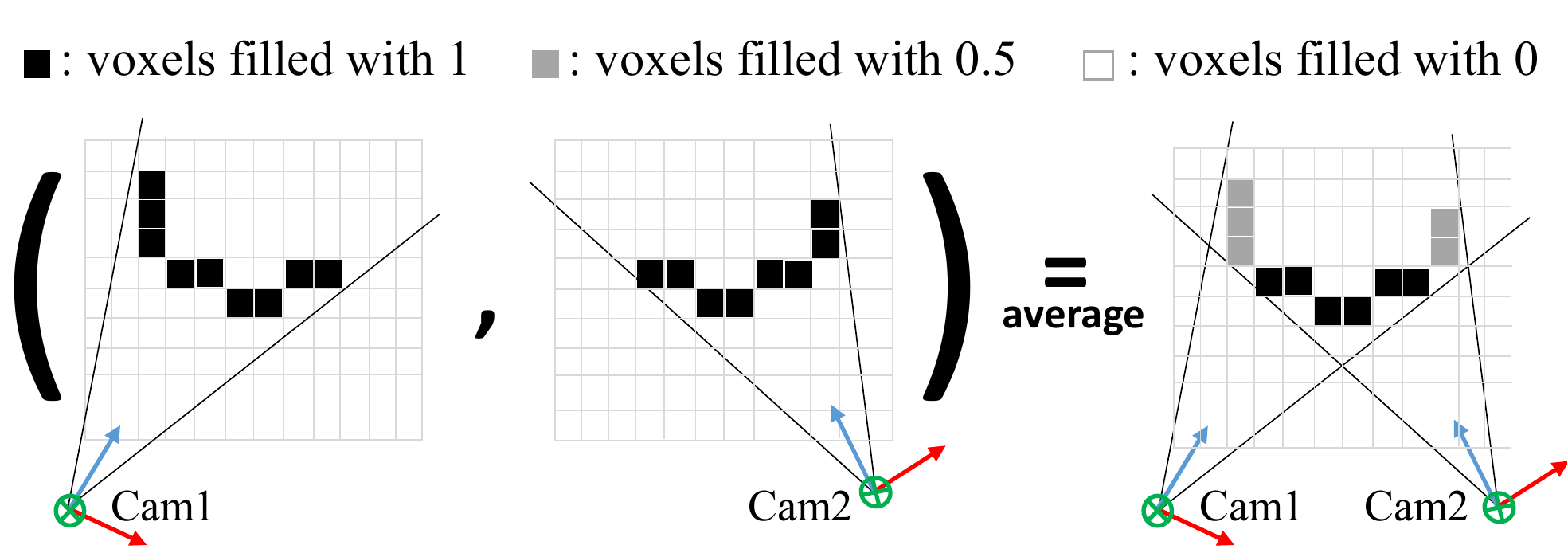}
\caption{\textbf{Top-view illustration of the volume fusion using two depth maps.}}
\label{fig:fig_backproj}
\end{figure}

\subsection{Volumetric Depth Fusion}
\label{subsec:Volumetric Depth Fusion}

In the previous section, we describe the first stage of our method specialized in regressing a per-view depth map~$\tilde{\mathcal{Z}}_\text{n}$ and an overlapping mask $\tilde{\mathcal{M}}_\text{n}$ to obtain a masked depth map~$\hat{\mathcal{Z}}_\text{n}$. To integrate these local estimations~$\{ \hat{\mathcal{Z}}_\text{n} \}_{\text{n}\text{=}1}^{N}$, we design a volume fusion that embeds a set of masked depth maps~$\{ \hat{\mathcal{Z}}_\text{n} \}_{\text{n}\text{=}1}^{N}$ into a unified scene volume~$\mathcal{V}^{U} {\in} \mathbb{R}^{(C\text{+}1) {\times} V_\text{x} {\times} V_\text{y} {\times} V_\text{z}}$. This stage aims at regressing the TSDF of a global scene by fusing local geometric information,~\ie,~$\{ \hat{\mathcal{Z}}_\text{n} \}_{\text{n}\text{=}1}^{N}$.

The traditional fusion process~\cite{colmap,galliani2015massively} attempts to achieve this reconstruction by using the photometric consistency check of the pixel values that are back-projected using the inferred depth maps~\cite{galliani2015massively,openmvg}.
However, these approaches suffer from the following drawbacks: 1)~the quality of the final reconstruction depends on the accuracy of the initial depth maps, and 2)~the brightness consistency assumption is violated under challenging conditions such as changing lighting conditions and homogeneously textured regions. Alternatively, recent learning-based approaches have been developed, such as RoutedFusion~\cite{routedfusion}. This approach produces high-quality 3D reconstruction but also requires reliable depth maps from RGB-D sensors. However, it is hardly suited for depth maps obtained via multi-view images which tend to be significantly more noisy.

To overcome these issues, our depth fusion strategy propagates the masked depth maps~$\{ \hat{\mathcal{Z}}_\text{n} \}_{\text{n}\text{=}1}^{N}$ as well as the feature volumes~$\mathcal{V}^{\mathcal{F}}$, which are computed from the image features $\mathcal{F}_\text{n}$ by back-projecting the image features into the world-referential coordinate system. This strategy allows the network to re-profile the surface of the 3D scene by computing the matching cost of the image features guided by the masked depth maps. To fuse the \text{per-view} masked depth maps~$\hat{\mathcal{Z}}$, we iteratively compute the per-view masked depth map and image feature. First, we declare a 3D volume following~\cite{atlas} and initialize all voxels with $0$. Then, we back-project a masked depth map and compute their voxel location $[i,j,k]^{\top}$. The value of each voxel occupied by a back-projected depth map is incremented by $1$. We repeat this process for all views, and then average the volume as shown in~\Fref{fig:fig_backproj}. This strategy allows to compute the 3D surface probability using all the previously computed depths.

The embedded voxels in the unified scene volume~$\mathcal{V}^{U}$ are the initial guidance for shaping the global structure of the target scene through volume fusion. To further enhance the geometric property of each embedded image feature, we introduce the concept of the Posed Convolution Layer (PosedConv) for an accurate reconstruction of the target scene.

\subsection{Posed Convolution Layer}
\label{subsec:Posed Convolution Layer}
Renowned stereo matching approaches~\cite{gcnet,psmnet,stereo_object,stereo_lidar} employ a series of 3D convolution layers (3DConv) to find dense correspondences between the left-right image features. Since this strategy proved to be effective, it became the most commonly used technique to build a cost volume. As a result, it has also been widely applied in un-rectified multi-view stereo pipelines~\cite{gpmvs,dpsnet,deepvideomvs}. Nonetheless, it appears that this representation is not appropriate for multi-view stereo. To understand this problem, we need to analyze both configurations: calibrated stereo pair and un-calibrated multi-view stereo. For calibrated stereo images, corresponding patches in both views are acquired from the same orientation (aligned optical axis); hence, the translation-invariant convolution operation is suitable to find consistent matches (see~\Fref{fig:fig_posed_conv}-(a)). However, MVS scenarios are more complex since images can be acquired from various orientations; therefore, the 3DConv is not appropriate since it is not rotation invariant (see~\Fref{fig:fig_posed_conv}-(b)). This phenomenon leads to poor matching quality when the orientations of the different viewpoints vary too greatly.

To cope with this limitation, we design rotation-dependent 3D convolution kernels, called PosedConv, using the known rotation matrix as shown in~\Fref{fig:fig_posed_conv}. First, we declare that the variable for our posed convolution layer which is named reservoir kernel $\mathbf{W}\in \mathbb{R}^{C_\text{out} \times C_\text{in} \times w \times w \times w}$ is similar to that of naive 3DConv. The parameters $C_\text{out}$ and $C_\text{in}$ denote the number of output and input channels, respectively, and the odd number $w$ represents the size of the 3D spatial window. Assuming that the reservoir kernel is aligned with the unified scene volume $\mathcal{V}^{U}$, we compute the rotated kernel $\mathbf{W}^{R}_\text{n}$ to be aligned to each corresponding camera viewpoint using the rotation matrix $\mathbf{R}_{1 \rightarrow \text{n}}$\footnote{We set the first camera coordinate as the world coordinate.}. The rotated kernel extracts more consistent features than the naive 3DConv since the proposed kernel is more robust to maintain the same receptive field even under rotation changes. Consequently, a pose-invariant feature volume~$\mathcal{V}^{R}_\text{n}$ are computed by convolving the rotated kernels $\mathbf{W}^{R}_\text{n}$ with the feature volume~$\mathcal{V}^{\mathcal{F}}_\text{n}$ as:
\begin{equation}
\begin{gathered}
\begin{split}
    \mathcal{V}^{R}_\text{n} 
    & = (\mathcal{V}^{\mathcal{F}}_\text{n} * \mathbf{W}^R_\text{n})(\mathbf{v}) \\
    & = \sum_{\mathbf{v}' \in \Omega } \mathcal{V}^{\mathcal{F}}_\text{n}(\mathbf{v}+\mathbf{v}') \cdot \mathbf{W}^R_\text{n}(\mathbf{v}') \\
    & = \sum_{\mathbf{v}' \in \Omega } \mathcal{V}^{\mathcal{F}}_\text{n}(\mathbf{v}+\mathbf{v}') \cdot \mathbf{W}(D_{R} \cdot \mathbf{v}'), \\
\end{split}
\end{gathered}
\end{equation}
where $\mathbf{v}\text{=}[i, j, k]^\intercal$ is the voxel coordinates in world referential, and $\Omega \text{=} \big\{ \mathbf{v}' {\in} \mathbb{Z}^{3} | [\text{-}w', \text{-}w', \text{-}w']^\intercal, ..., [\text{+}w', \text{+}w', \text{+}w']^\intercal\big\}$ is a set of signed distances from the center of the reservoir kernel to each voxel $\mathbf{v}$ in the kernel where $w'\text{=}(w\text{-}1)/2$.
The dot operator $\cdot$ indicates dot product,
and $D_{R}$ represents the modified rotation matrix through our \emph{Discrete Kernel Rotation}\footnote{The details of $D_{R}$ is available in the supplementary material.}. 
Then, we calculate the part of the unified scene volume $\mathcal{V}^{U}$ by averaging a set of the pose-invariant feature volume $\{ \mathcal{V}^{R}_\text{n} \}_{\text{n}\text{=}1}^{N}$.
Our PosedConv, elaborately designed for varying camera orientation, plays an important role in building the unified scene volume $\mathcal{V}^{U}$ and robust 3D scene reconstruction.

\subsection{Volumetric 3D Reconstruction}
\label{subsec:Volumetric 3D Reconstruction}
The unified scene volume $\mathcal{V}^{U}$ is aggregated through stacked hourglass 3DConv layers~\cite{associate_embedding,hourglass,psmnet} to compute the TSDF of the entire scene. After the aggregation process, we obtain a TSDF volume $\tilde{\mathcal{V}}^{\text{TSDF}} \in \mathbb{R}^{V_\text{x} {\times} V_\text{y} {\times} V_\text{z}}$, as shown in~\Fref{fig:fig_arch}.
The estimated TSDF volume $\tilde{\mathcal{V}}^{\text{TSDF}}$ involves the value of truncated signed distance from the surface, and it is trained in a supervised manner as follows:
\begin{equation}
\begin{gathered}
    \mathcal{L}_{\text{TSDF}} = \sum_{(x,y,z)} \left | \tilde{\mathcal{V}}^{\text{TSDF}}_{x,y,z}  - {\mathcal{V}}^{\text{TSDF}}_{x,y,z} \right |_{1},
\label{eq:TSDF-loss}
\end{gathered}
\end{equation}
where ${\mathcal{V}}^{\text{TSDF}}_{x,y,z}$ is the ground-truth TSDF volume and $\left | \cdot \right |_{1}$ is the absolute distance measurements (\ie L1-loss), and $(x,y,z)$ is the voxel location within the TSDF volume.

Finally, our network is trained in an end-to-end manner with the following three different losses:
\begin{equation}
    \mathcal{L}_{tot} = \alpha \mathcal{L}_{\mathcal{Z}} + \beta \mathcal{L}_{\mathcal{M}} + \gamma \mathcal{L}_{\text{TSDF}},
\end{equation}
where $\alpha$, $\beta$, and $\gamma$ are $1.0$, $0.5$, and $2.0$, respectively. 
The depth loss $\mathcal{L}_{\mathcal{Z}}$ and the overlapping loss $\mathcal{L}_{\mathcal{M}}$ guide the network to perform local multi-view stereo matching for explicit depth estimation. The TSDF loss $ \mathcal{L}_{\text{TSDF}}$ is intended to transform the explicit geometry depth maps and the per-view image features into an implicit representation through our volume fusion.

\begin{table*}[!t]
\centering
\begin{tabular}{lccccccccc}
\Xhline{2\arrayrulewidth}
\multirow{2}{*}{Method}      & \multicolumn{4}{c}{2D Depth Evaluation}     & & \multicolumn{4}{c}{3D Geometry Evaluation} \\ \cline{2-5} \cline{7-10}
                & AbsRel & AbsDiff & SqRel & RMSE             & & $\mathcal{L}_{1}$ & Acc & Comp & F-score \\ \Xhline{2\arrayrulewidth}
COLMAP~\cite{colmap}         & .137   & .264    & .138  & .502 &            & .599 & .069 & .135 & \textbf{.558}\\ 
MVDepthNet~\cite{mvdepthnet} & .098   & .191    & .061  & .293 &            & .518 & .040 & .240 & .329 \\ 
GPMVS~\cite{gpmvs}           & .130   & .239    & .339  & .472 &            & .475 & \textbf{.031} & .879 & .304 \\ 
DPSNet~\cite{dpsnet}         & .087   & .158    & .035  & .232 &            & .421 & .045 & .284 & .344 \\ 
Murez~\etal~\cite{atlas}     & .061   & .120    & .042  & .248 &            & .162 & .065 & .130 & .499 \\ 


VolumeFusion (ours) & \textbf{.049} & \textbf{.084} & \textbf{.021} & \textbf{.164} & & \textbf{.141} & .038 & \textbf{.125} & .508  \\ \Xhline{2\arrayrulewidth}
\end{tabular}
\vspace{1mm}
\caption{\textbf{Quantitative results on ScanNet dataset~\cite{scannet}.} We provide two metrics: depth evaluation and 3D geometry evaluation.}
\label{table:scannet}
\vspace{1mm}
\end{table*}

\section{Experiment}
\label{sec:Experiment}

\subsection{Implementation Details and Dataset}
\label{subsec:Dataset and implementation}

Following Murez~\etal~\cite{atlas}, we conduct the assessment of our method on the ScanNet dataset~\cite{scannet}. The ScanNet dataset consists of 800 indoor scenes. Each scene contains a sequence of RGB images, the corresponding ground-truth depth maps, and the cameras' parameters. Among them, 700 scenes are used for training while the remaining 100 scenes constitute our testing set. To obtain the ground-truth TSDF volume $\mathcal{V}^{\text{TSDF}}$, we follow the original scheme proposed by a previous study~\cite{atlas}.

In the first stage of our network, we use three local views per frame to operate the local multi-view stereo matching. The input size of the image is $480\text{(H)} {\times} 640\text{(W)}$ and the resolution of the estimated depth is $120\text{(H)} {\times} 160\text{(W)}$. The size of the overlapping mask is identical to that of the depth map. After PosedConv extracts pose-invariant scene volume 
$\mathcal{V}^{R}_\text{n}$, we integrate the per-view information 
($\{ \hat{\mathcal{Z}}_{\text{n}} \}_{\text{n}\text{=}1}^{N}$ and
$\{ \mathcal{V}^{R}_\text{n} \}_{\text{n}\text{=}1}^{N}$) into the unified scene volume $\mathcal{V}^{U}$. Finally, the second stage of our network is trained in a supervised manner via the TSDF loss $\mathcal{L}_{\text{TSDF}}$.
The unified scene volume $\mathcal{V}^{U}$ and the TSDF volume $\mathcal{V}^{\text{TSDF}}$ covers $0.04m$ per voxel. The resolution of the unified scene volume $\mathcal{V}^{U} {\in} \mathbb{R}^{(C\text{+}1) {\times} V_\text{x} {\times} V_\text{y} {\times} V_\text{z}}$ is 160($V_\text{x}$)$\times$64($V_\text{y}$)$\times$160($V_\text{z}$) for training and 416($V_\text{x}$)$\times$128($V_\text{y}$)$\times$416($V_\text{z}$). We set the initial learning rate as $0.0001$ and drop the learning rate by half after 50 epochs. We train our network for 100 epochs with 8 NVIDIA RTX 3090 GPUs, which takes about two days.

\subsection{Comparison with State-of-the-art Methods}
\label{subsec:Comparison with state-of-the-art methods}
To validate the effectiveness of the proposed approach, we compare the reconstruction performance to a variety of traditional geometry-based and deep learning-based methods~\cite{colmap,gpmvs,dpsnet,mvdepthnet,atlas} in both 3D space and the 2.5D depth map domain. Specifically, we use the four common quantitative measures (AbsRel, AbsDiff, SqRel, and RMSE)$^{\ref{ft:metric}}$ of depth map quality and three common criteria ($\mathcal{L}_{1}$, Accuracy(Acc), Completeness(Comp), and F-score)\footnote{\label{ft:metric} We provide a detailed description of these metrics in our supplementary material} on 3D reconstruction quality. The quantitative results are reported in~\Tref{table:scannet}. The evaluation is conducted with the 100 scenes from the ScanNet~\cite{scannet} test set following the evaluation pipeline described in Murez~\etal~\cite{atlas}. As our experimental results show, the proposed method outperforms all competitors for all evaluation metrics of depth map by a large margin. We conjecture that the significant performance gap results from our depth map fusion method, which effectively matches multiple images even at a large viewpoint difference and increases the number of observations for matching.

In the evaluation of 3D reconstruction in~\Tref{table:scannet}, our approach outperforms the others especially on the $\mathcal{L}_{1}$ and Comp. It suggests that our method is outstanding for shaping the global structure of the scene. For the Acc metric, our method demonstrates the second-best results after GPMVS~\cite{gpmvs}, which shows that temporal fusion~\cite{gpmvs} is a potential method to improve the quality of the multi-view depth map. However, for the 3D reconstruction of the overall scene ($\mathcal{L}_{1}$, Comp), our method largely outperforms the temporal fusion method~\cite{gpmvs}. The best F-score is obtained by COLMAP~\cite{colmap}. As described in DPSNet~\cite{dpsnet}, COLMAP~\cite{colmap} has the strength to accurately reconstruct the edges or corners where distinctive features are extracted.

\begin{figure*}[!t]
\centering
\includegraphics[width=1.00\linewidth]{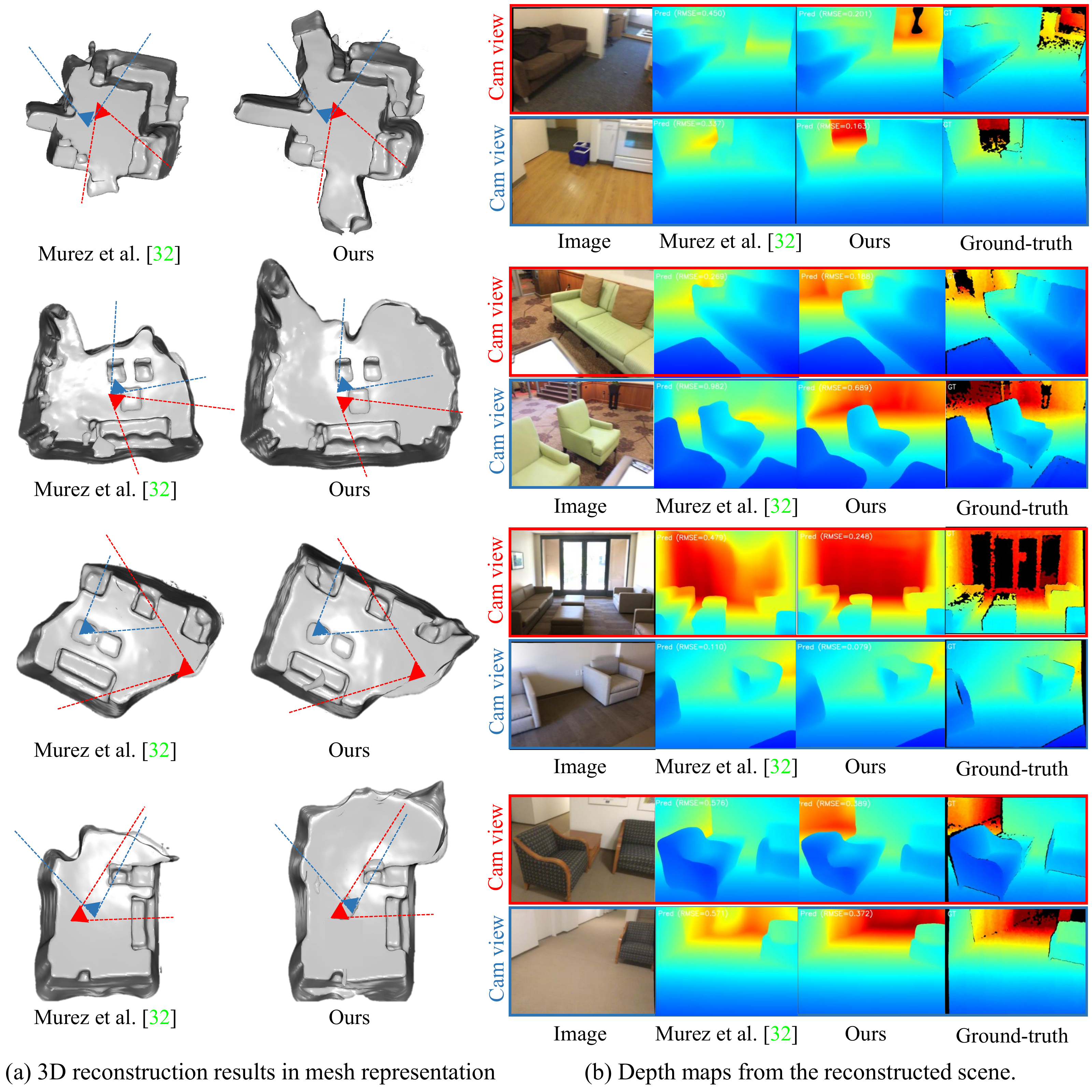}
\caption{\textbf{Qualitative reconstruction results in ScanNet dataset~\cite{scannet}.} 
(a) 3D scene reconstruction results from the recent work (Murez~\etal~\cite{atlas}) and ours. 
(b) Depth results from the two selected camera viewpoints. Overall, our method shows better results, especially for the global structure of the target scenes.}   
\label{fig:fig_qual}
\end{figure*}

\begin{figure*}[!t]
\centering
\includegraphics[width=1.00\linewidth]{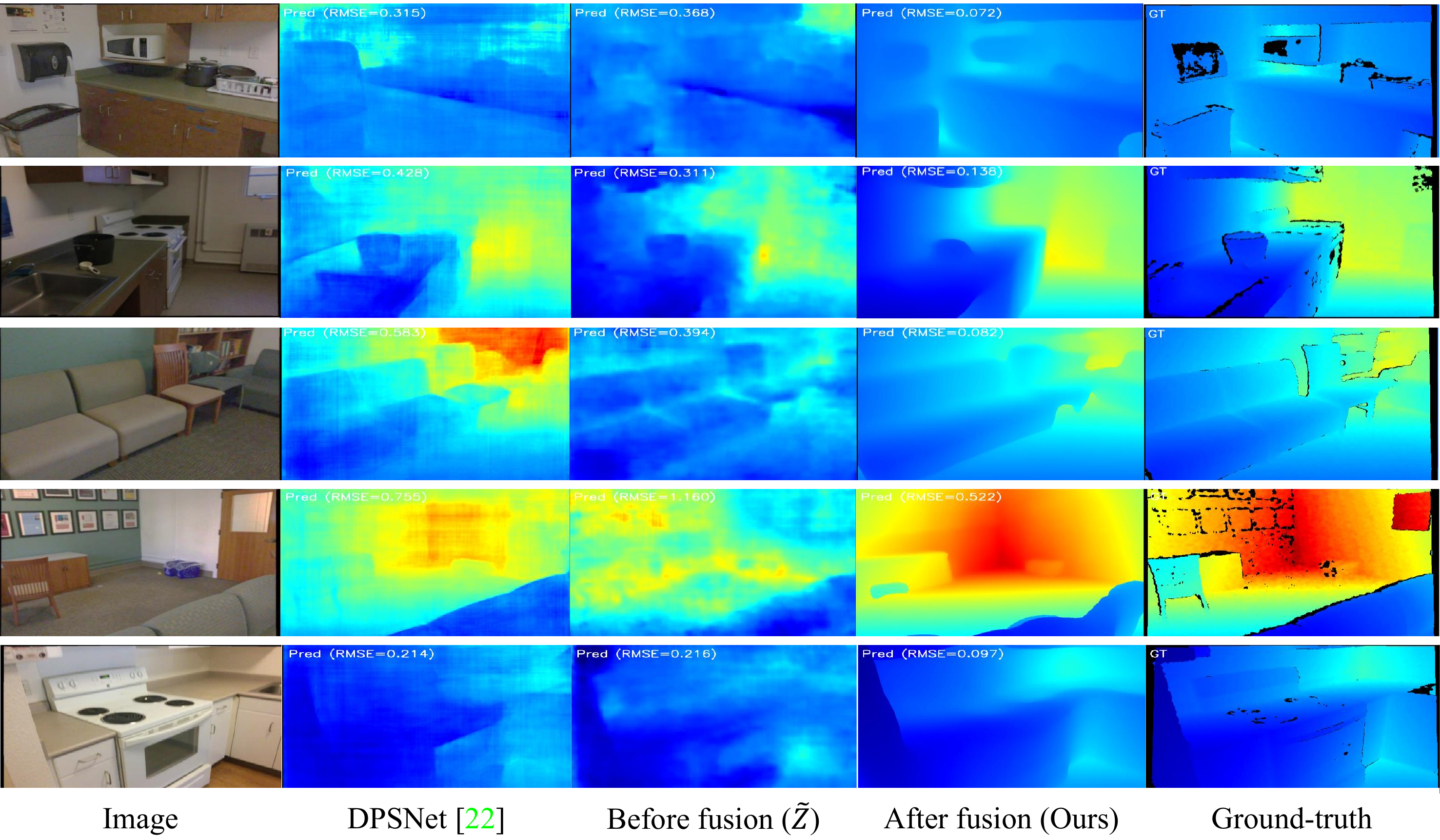}
\caption{\textbf{Qualitative results of our depth map fusion and recent depth-based MVS method~\cite{dpsnet}.} 
Compared to our depth estimation from the local multi-view stereo (Before fusion $\tilde{\mathcal{Z}}$) and from the final estimation (After fusion), it demonstrates that our depth fusion scheme is trained to re-profile the 3D surface so that the quality of depth maps after fusion outperforms depth maps before fusion $\tilde{\mathcal{Z}}$.}   
\label{fig:fig_depth}
\end{figure*}

Additionally, we show qualitative results of the depth fusion results in~\Fref{fig:fig_depth}, and the 3D scene reconstruction in~\Fref{fig:fig_qual}.
Compared to the most recent competitive technique~\cite{atlas}, our method better preserves the global structure of the scene, especially for the shapes of complex rooms -- \eg with corridors. Moreover, concerning the quality of depth estimation, our method shows improved depth accuracy after fusion through volume fusion. 
We attribute the performance improvement to our two-stage approach that exploits the pose-invariant features that re-profile the surface of the scene with the robust matching in a world-referential coordinate, \ie, the unified scene volume $\mathcal{V}^{U}$).


\begin{table}[!t]
\centering
\resizebox{1.00\linewidth}{!}{
\begin{tabular}{lcccc}
\Xhline{2\arrayrulewidth}
\multirow{2}{*}{Method \ \ \ \ \ \ \ } & \multicolumn{4}{c}{Evaluation} \\ \cline{2-5}
& AbsRel & RMSE & $\mathcal{L}_{1}$ & F-score \\ \Xhline{2\arrayrulewidth}

3D Conv & .058 & .231 & .166 & .460 \\


PosedConv & \textbf{.049} & \textbf{.164} & \textbf{.141} & \textbf{.508} \\ \Xhline{2\arrayrulewidth}

\end{tabular}
}
\caption{\textbf{Ablation study of PosedConv.} We compare original convolution layer (3D Conv), and our PosedConv.}
\label{table:ablation-posedconv}
\vspace{-2mm}
\end{table}
\begin{table}[!t]
\centering
\resizebox{1.00\linewidth}{!}{
\begin{tabular}{lcccccc}
\Xhline{2\arrayrulewidth}
\multirow{2}{*}{Method}   & \multicolumn{2}{c}{Preserve}     & & \multicolumn{2}{c}{Evaluation} \\ \cline{2-3}\cline{5-6}
                          & Posed  & Depth  & & \multirow{2}{*}{RMSE} & \multirow{2}{*}{$\mathcal{L}_{1}$} \\ 
                          & Conv   & Fusion & & \\ \Xhline{2\arrayrulewidth}

Single-stage~\cite{atlas} &        &        & & .248 & .162 \\
w/o depth        & \cmark &        & & .236 & .159 \\ 

Two-stage (ours)          & \cmark & \cmark & & \textbf{.164} & \textbf{.141} \\ \Xhline{2\arrayrulewidth}

\end{tabular}
}
\caption{\textbf{Ablation study for depth fusion.} Note that $\cmark$ means preserve the depth fusion process as shown in~\Fref{fig:fig_arch}.}
\label{table:ablation-depthfusion}
\vspace{-2mm}
\end{table}
\begin{table}[!t]
\centering
\resizebox{1.00\linewidth}{!}{
\begin{tabular}{lcccc}
\Xhline{2\arrayrulewidth}
\multirow{2}{*}{Method} & \multicolumn{4}{c}{Evaluation} \\ \cline{2-5}
& AbsRel & RMSE & $\mathcal{L}_{1}$ & F-score \\ \Xhline{2\arrayrulewidth}

Ours w/o ${\mathcal{M}}$ & .060 & .238 & .162 & .475 \\

Ours w/ ${\mathcal{M}}$ & \textbf{.049} & \textbf{.164} & \textbf{.141} & \textbf{.508} \\ \Xhline{2\arrayrulewidth}
\end{tabular}
}
\caption{\textbf{Ablation study of an overlapping mask ${\mathcal{M}}$.}}
\label{table:ablation-mask}
\vspace{-2mm}
\end{table}

\subsection{Ablation Study}
\label{subsec:Ablation study}
In this section, we propose to evaluate the contribution of each proposed component (PosedConv, depth map fusion, and overlapping mask), through an extensive ablation study. The obtained results are shown in Tables~\textcolor{red}{2},~\textcolor{red}{3}, and~\textcolor{red}{4}.

In~\Tref{table:ablation-posedconv}, we validate the PosedConv by comparing with the origin 3D~Conv as in~\Fref{fig:fig_posed_conv}-(\textcolor{red}{a})). This result highlights the relevance of PosedConv since it consistently improves the depth (AbsRel and RMSE) and reconstruction ($\mathcal{L}_{1}$ and F-score) accuracy. We attribute these results to the rotation-invariant and translation-invariant features extracted from our PosedConv, which helps to re-profile the 3D surface in the depth map fusion stage (\Fref{fig:fig_arch}).

In~\Tref{table:ablation-depthfusion}, we conduct an ablation study regarding the two-stage strategy of our method. Single-stage represents the direct TSDF regression as proposed in Murez~\etal~\cite{atlas}, and two-stage means our volume fusion network equipped with the PosedConv and depth map fusion.
To verify the necessity of the depth fusion (\ie, depth maps embedding), we also include one additional method (w/o depth embedding) whose unified scene volume $\mathcal{V}^{U}$ only contains pose-invariant feature volume $\mathcal{V}^{R}$. The results demonstrate that our two-stage approach outperforms the single-stage strategy~\cite{atlas}. Moreover, when we embed masked depth maps $\hat{\mathcal{M}}$, the performance gap to the previous work~\cite{atlas} increases. Thanks to our volume fusion network with PosedConv and differentiable depth map fusion scheme, we obtain the best 3D reconstruction results.

Lastly, we validate the overlapping mask $\mathcal{M}$ in~\Tref{table:ablation-mask}. This mask is used to filter out the depth values at the non-overlapping region between local multi-view images. It shows that applying overlapping masks consistently improves the quality of the depth maps and 3D reconstruction. 
Based on these results, we confirm that using overlapping masks under varying camera motion is effective for depth map fusion.


\section{Conclusion}
\label{sec:conclusion}
In this work, we presented an end-to-end volume fusion network for 3D scene reconstruction using a set of images with known camera poses. The specificity of our strategy is its two-stage structure that mimics traditional techniques: local multi-view stereo and Volumetric Depth Fusion. For the local depth maps estimation, we designed a novel multi-view stereo method that also estimates an overlapping mask. This additional output enables filtering out of the depth measurements that do not have the pixel correspondence among the neighbor views, which in turn, improves the overall reconstruction of the scene. In our depth fusion, we fuse the masked depth maps as well as the image features to re-profile the 3D surface by computing the matching cost. To improve the robustness of matching between images with different orientations -- in the world-referential coordinate (\ie, unified scene volume), we introduce the Posed Convolution Layer that extracts pose-invariant features. Finally, our network infers a TSDF volume that describes the global structure of the target scene. Despite the consistency and accuracy of our 3D reconstructions, volumetric representation requires huge computation power and memory that restricts the resolution of the resulting TSDF.
To cope with this problem, future works for volume-free fusion via local multi-view information should be explored. In this context, our method represents a solid base for the development of future research in multi-view stereo and depth map fusion for 3D scene reconstruction.

\section*{ACKNOWLEDGMENT}
This work was supported by NAVER LABS Corporation [SSIM: Semantic and scalable indoor mapping]


{\small
\bibliographystyle{ieee_fullname}
\bibliography{egbib}

\begin{thebibliography}{10}\itemsep=-1pt

\bibitem{trans_fusion}
Alja{\v{z}} Bo{\v{z}}i{\v{c}}, Pablo Palafox, Justus Thies, Angela Dai, and
  Matthias Nie{\ss}ner.
\newblock Transformerfusion: Monocular rgb scene reconstruction using
  transformers.
\newblock {\em arXiv preprint arXiv:2107.02191}, 2021.

\bibitem{campbell2008using}
Neill~DF Campbell, George Vogiatzis, Carlos Hern{\'a}ndez, and Roberto Cipolla.
\newblock Using multiple hypotheses to improve depth-maps for multi-view
  stereo.
\newblock In {\em European Conference on Computer Vision}, pages 766--779.
  Springer, 2008.

\bibitem{psmnet}
Jia-Ren Chang and Yong-Sheng Chen.
\newblock Pyramid stereo matching network.
\newblock In {\em Proceedings of the IEEE Conference on Computer Vision and
  Pattern Recognition}, pages 5410--5418, 2018.

\bibitem{point_mvs}
Rui Chen, Songfang Han, Jing Xu, and Hao Su.
\newblock Point-based multi-view stereo network.
\newblock In {\em Proceedings of the IEEE International Conference on Computer
  Vision}, pages 1538--1547, 2019.

\bibitem{stereo_lidar}
Jaesung Choe, Kyungdon Joo, Tooba Imtiaz, and In~So Kweon.
\newblock Volumetric propagation network: Stereo-lidar fusion for long-range
  depth estimation.
\newblock {\em IEEE Robotics and Automation Letters}, 6(3):4672--4679, 2021.

\bibitem{stereo_object}
Jaesung Choe, Kyungdon Joo, Francois Rameau, and In~So Kweon.
\newblock Stereo object matching network.
\newblock {\em arXiv preprint arXiv:2103.12498}, 2021.

\bibitem{plane_sweep_00}
Robert~T Collins.
\newblock A space-sweep approach to true multi-image matching.
\newblock In {\em Proceedings CVPR IEEE Computer Society Conference on Computer
  Vision and Pattern Recognition}, pages 358--363. IEEE, 1996.

\bibitem{curless1996volumetric}
Brian Curless and Marc Levoy.
\newblock A volumetric method for building complex models from range images.
\newblock In {\em Proceedings of the 23rd annual conference on Computer
  graphics and interactive techniques}, 1996.

\bibitem{scannet}
Angela Dai, Angel~X Chang, Manolis Savva, Maciej Halber, Thomas Funkhouser, and
  Matthias Nie{\ss}ner.
\newblock Scannet: Richly-annotated 3d reconstructions of indoor scenes.
\newblock In {\em Proceedings of the IEEE Conference on Computer Vision and
  Pattern Recognition}, pages 5828--5839, 2017.

\bibitem{deepvideomvs}
Arda D{\"u}z{\c{c}}eker, Silvano Galliani, Christoph Vogel, Pablo Speciale,
  Mihai Dusmanu, and Marc Pollefeys.
\newblock Deepvideomvs: Multi-view stereo on video with recurrent
  spatio-temporal fusion.
\newblock In {\em IEEE Conf. Comput. Vis. Pattern Recog.}, 2021.

\bibitem{esteban2004silhouette}
Carlos~Hern{\'a}ndez Esteban and Francis Schmitt.
\newblock Silhouette and stereo fusion for 3d object modeling.
\newblock {\em Computer Vision and Image Understanding}, 96(3):367--392, 2004.

\bibitem{intro_mvs}
Yasutaka Furukawa and Carlos Hern{\'a}ndez.
\newblock Multi-view stereo: A tutorial.
\newblock {\em Foundations and Trends{\textregistered} in Computer Graphics and
  Vision}, 9(1-2):1--148, 2015.

\bibitem{furukawa2009accurate}
Yasutaka Furukawa and Jean Ponce.
\newblock Accurate, dense, and robust multiview stereopsis.
\newblock {\em IEEE transactions on pattern analysis and machine intelligence},
  32(8):1362--1376, 2009.

\bibitem{galliani2015massively}
Silvano Galliani, Katrin Lasinger, and Konrad Schindler.
\newblock Massively parallel multiview stereopsis by surface normal diffusion.
\newblock In {\em Proceedings of the IEEE International Conference on Computer
  Vision}, pages 873--881, 2015.

\bibitem{gallup2007real}
David Gallup, Jan-Michael Frahm, Philippos Mordohai, Qingxiong Yang, and Marc
  Pollefeys.
\newblock Real-time plane-sweeping stereo with multiple sweeping directions.
\newblock In {\em 2007 IEEE Conference on Computer Vision and Pattern
  Recognition}, pages 1--8. IEEE, 2007.

\bibitem{cascade_mvs}
Xiaodong Gu, Zhiwen Fan, Siyu Zhu, Zuozhuo Dai, Feitong Tan, and Ping Tan.
\newblock Cascade cost volume for high-resolution multi-view stereo and stereo
  matching.
\newblock In {\em IEEE Conf. Comput. Vis. Pattern Recog.}, 2020.

\bibitem{plane_sweep_03}
Hyowon Ha, Sunghoon Im, Jaesik Park, Hae-Gon Jeon, and In~So Kweon.
\newblock High-quality depth from uncalibrated small motion clip.
\newblock In {\em Proceedings of the IEEE conference on computer vision and
  pattern Recognition}, pages 5413--5421, 2016.

\bibitem{geometry}
Richard Hartley and Andrew Zisserman.
\newblock {\em Multiple view geometry in computer vision}.
\newblock Cambridge university press, 2003.

\bibitem{hernandez2007probabilistic}
Carlos Hern{\'a}ndez, George Vogiatzis, and Roberto Cipolla.
\newblock Probabilistic visibility for multi-view stereo.
\newblock In {\em 2007 IEEE Conference on Computer Vision and Pattern
  Recognition}, pages 1--8. IEEE, 2007.

\bibitem{gpmvs}
Yuxin Hou, Juho Kannala, and Arno Solin.
\newblock Multi-view stereo by temporal nonparametric fusion.
\newblock In {\em Proceedings of the IEEE International Conference on Computer
  Vision}, pages 2651--2660, 2019.

\bibitem{plane_sweep_02}
Sunghoon Im, Hyowon Ha, Gyeongmin Choe, Hae-Gon Jeon, Kyungdon Joo, and In~So
  Kweon.
\newblock Accurate 3d reconstruction from small motion clip for rolling shutter
  cameras.
\newblock {\em IEEE transactions on pattern analysis and machine intelligence},
  41(4):775--787, 2018.

\bibitem{dpsnet}
Sunghoon Im, Hae-Gon Jeon, Stephen Lin, and In~So Kweon.
\newblock Dpsnet: end-to-end deep plane sweep stereo.
\newblock In {\em International Conference on Learning Representations (ICLR)},
  2019.

\bibitem{izadi2011kinectfusion}
Shahram Izadi, Richard~A Newcombe, David Kim, Otmar Hilliges, David Molyneaux,
  Steve Hodges, Pushmeet Kohli, Jamie Shotton, Andrew~J Davison, and Andrew
  Fitzgibbon.
\newblock Kinectfusion: real-time dynamic 3d surface reconstruction and
  interaction.
\newblock In {\em ACM SIGGRAPH 2011 Talks}, 2011.

\bibitem{ji2017surfacenet}
Mengqi Ji, Juergen Gall, Haitian Zheng, Yebin Liu, and Lu Fang.
\newblock Surfacenet: An end-to-end 3d neural network for multiview stereopsis.
\newblock In {\em Proceedings of the IEEE International Conference on Computer
  Vision}, pages 2307--2315, 2017.

\bibitem{kahler2015hierarchical}
Olaf K{\"a}hler, Victor Prisacariu, Julien Valentin, and David Murray.
\newblock Hierarchical voxel block hashing for efficient integration of depth
  images.
\newblock {\em IEEE Robotics and Automation Letters}, 1(1):192--197, 2015.

\bibitem{gcnet}
Alex Kendall, Hayk Martirosyan, Saumitro Dasgupta, and Peter Henry.
\newblock End-to-end learning of geometry and context for deep stereo
  regression.
\newblock In {\em 2017 IEEE International Conference on Computer Vision
  (ICCV)}, pages 66--75. IEEE, 2017.

\bibitem{normal_assited}
Uday Kusupati, Shuo Cheng, Rui Chen, and Hao Su.
\newblock Normal assisted stereo depth estimation.
\newblock In {\em IEEE Conf. Comput. Vis. Pattern Recog.}, 2020.

\bibitem{old_mvs_point_cloud_00}
Maxime Lhuillier and Long Quan.
\newblock A quasi-dense approach to surface reconstruction from uncalibrated
  images.
\newblock {\em IEEE transactions on pattern analysis and machine intelligence},
  27(3):418--433, 2005.

\bibitem{lorensen1987marching}
William~E Lorensen and Harvey~E Cline.
\newblock Marching cubes: A high resolution 3d surface construction algorithm.
\newblock {\em ACM siggraph computer graphics}, 21(4):163--169, 1987.

\bibitem{kitti-stereo}
Moritz Menze and Andreas Geiger.
\newblock Object scene flow for autonomous vehicles.
\newblock In {\em Proceedings of the IEEE Conference on Computer Vision and
  Pattern Recognition}, pages 3061--3070, 2015.

\bibitem{openmvg}
Pierre Moulon, Pascal Monasse, Romuald Perrot, and Renaud Marlet.
\newblock Openmvg: Open multiple view geometry.
\newblock In {\em International Workshop on Reproducible Research in Pattern
  Recognition}, pages 60--74. Springer, 2016.

\bibitem{atlas}
Zak Murez, Tarrence van As, James Bartolozzi, Ayan Sinha, Vijay Badrinarayanan,
  and Andrew Rabinovich.
\newblock Atlas: End-to-end 3d scene reconstruction from posed images.
\newblock In {\em Eur. Conf. Comput. Vis.}, 2020.

\bibitem{associate_embedding}
Alejandro Newell, Zhiao Huang, and Jia Deng.
\newblock Associative embedding: End-to-end learning for joint detection and
  grouping.
\newblock In {\em Advances in Neural Information Processing Systems}, pages
  2277--2287, 2017.

\bibitem{hourglass}
Alejandro Newell, Kaiyu Yang, and Jia Deng.
\newblock Stacked hourglass networks for human pose estimation.
\newblock In {\em European Conference on Computer Vision}, pages 483--499.
  Springer, 2016.

\bibitem{niessner2013real}
Matthias Nie{\ss}ner, Michael Zollh{\"o}fer, Shahram Izadi, and Marc
  Stamminger.
\newblock Real-time 3d reconstruction at scale using voxel hashing.
\newblock {\em ACM Transactions on Graphics (ToG)}, 32(6):1--11, 2013.

\bibitem{paschalidou2018raynet}
Despoina Paschalidou, Osman Ulusoy, Carolin Schmitt, Luc Van~Gool, and Andreas
  Geiger.
\newblock Raynet: Learning volumetric 3d reconstruction with ray potentials.
\newblock In {\em Proceedings of the IEEE Conference on Computer Vision and
  Pattern Recognition}, pages 3897--3906, 2018.

\bibitem{colmap}
Johannes~L Sch{\"o}nberger, Enliang Zheng, Jan-Michael Frahm, and Marc
  Pollefeys.
\newblock Pixelwise view selection for unstructured multi-view stereo.
\newblock In {\em European Conference on Computer Vision}, pages 501--518.
  Springer, 2016.

\bibitem{neural_recon}
Jiaming Sun, Yiming Xie, Linghao Chen, Xiaowei Zhou, and Hujun Bao.
\newblock Neuralrecon: Real-time coherent 3d reconstruction from monocular
  video.
\newblock In {\em IEEE Conf. Comput. Vis. Pattern Recog.}, 2021.

\bibitem{mvdepthnet}
Kaixuan Wang and Shaojie Shen.
\newblock Mvdepthnet: Real-time multiview depth estimation neural network.
\newblock In {\em 2018 International Conference on 3D Vision (3DV)}, pages
  248--257. IEEE, 2018.

\bibitem{routedfusion}
Silvan Weder, Johannes Schonberger, Marc Pollefeys, and Martin~R Oswald.
\newblock Routedfusion: Learning real-time depth map fusion.
\newblock In {\em IEEE Conf. Comput. Vis. Pattern Recog.}, 2020.

\bibitem{neuralfusion}
Silvan Weder, Johannes~L Schonberger, Marc Pollefeys, and Martin~R Oswald.
\newblock Neuralfusion: Online depth fusion in latent space.
\newblock In {\em Proceedings of the IEEE/CVF Conference on Computer Vision and
  Pattern Recognition}, pages 3162--3172, 2021.

\bibitem{plane_sweep_01}
Ruigang Yang and Marc Pollefeys.
\newblock Multi-resolution real-time stereo on commodity graphics hardware.
\newblock In {\em 2003 IEEE Computer Society Conference on Computer Vision and
  Pattern Recognition, 2003. Proceedings.}, volume~1, pages I--I. IEEE, 2003.

\bibitem{mvsnet}
Yao Yao, Zixin Luo, Shiwei Li, Tian Fang, and Long Quan.
\newblock Mvsnet: Depth inference for unstructured multi-view stereo.
\newblock In {\em Proceedings of the European Conference on Computer Vision
  (ECCV)}, pages 767--783, 2018.

\bibitem{ganet}
Feihu Zhang, Victor Prisacariu, Ruigang Yang, and Philip~HS Torr.
\newblock Ga-net: Guided aggregation net for end-to-end stereo matching.
\newblock In {\em Proceedings of the IEEE Conference on Computer Vision and
  Pattern Recognition}, pages 185--194, 2019.

\end{thebibliography}
}

\newpage

\renewcommand*{\thesection}{\Alph{section}}

{\large \textbf{Supplementary Material}}

\setcounter{section}{0}

\section{Overview}
This document provides additional information that we did not fully cover in the main paper, \textbf{\textit{VolumeFusion: Deep Depth Fusion for 3D Scene Reconstruction}}. All references are consistent with the main paper.

\section{Evaluation Metrics}
In this section, we describe the metrics for depth and 3D reconstruction evaluation. Depth evaluation involves four different metrics: Abs~Rel, Abs~Diff, Sq~Rel, and RMSE. Each of these metrics is calculated as:

\begin{subequations}
\begin{align}
\text{Abs~Rel} &= \quad \frac{1}{n}\sum_{(u,v)} \left | z_{(u,v)} - \tilde{z}_{(u,v)} \right |_{1} / \tilde{z}_{(u,v)}, \\
\text{Abs~Diff} &= \quad \frac{1}{n}\sum_{(u,v)} \left | z_{(u,v)} - \tilde{z}_{(u,v)} \right |_{1}, \\
\text{Sq~Rel} &= \quad \frac{1}{n}\sum_{(u,v)} \left | z_{(u,v)} - \tilde{z}_{(u,v)} \right |^{2}  / \tilde{z}_{(u,v)}, \\
\text{RMSE} &= \quad \sqrt{\frac{1}{n}\sum_{(u,v)} \left | z_{(u,v)} - \tilde{z}_{(u,v)} \right |^{2}},
\end{align}
\end{subequations}
where $n$ is the number of pixels within both valid ground-truth and predictions, $z_{(u,v)}$ is the depth value at the pixel $(u, v)$ in the ground-truth depth map $z$, $\tilde{z}_{(u,v)}$ is the depth value at the pixel $(u, v)$ in the inferred depth map $\tilde{z}$, and $|\cdot|_{1}$ represents absolute distance.
Note that the inferred depth map of our method is the final fused depth results after the scene reconstruction.

Regarding the 3D geometry evaluation, we propose four different metrics:
$\mathcal{L}_{1}$, accuracy~(Acc), completeness~(Comp), and F-score.
$\mathcal{L}_{1}$ is the absolute difference between the ground-truth TSDF and the inferred TSDF, the accuracy is the distance from predicted points to the ground-truth points, completeness is the distance from the ground-truth points to the predicted points, and F-score is the harmonic mean of the precision and recall.
Each of these metrics is calculated as:
\begin{subequations}
\begin{align}
\mathcal{L}_{1} &= \quad \text{mean}_{a < 1} \left | a - \tilde{a} \right |_{1} \\
\text{Acc} &= \quad \text{mean}_{\tilde{\textbf{p}} \in \tilde{\mathcal{P}}} (\text{min}_{\textbf{p} \in \mathcal{P}} \left | \textbf{p} - \tilde{\textbf{p}} \right |_{1} )\\
\text{Comp} &= \quad \text{mean}_{\textbf{p} \in \mathcal{P}} (\text{min}_{\tilde{\textbf{p}} \in \tilde{\mathcal{P}}} \left | \textbf{p} - \tilde{\textbf{p}} \right |_{1} ) \\
\text{F-score} &= \quad \frac{2 \times \text{Prec} \times \text{Recall}}{\text{Prec} + \text{Recall}}
\end{align}
\vspace{+4mm}
\end{subequations}
where $\tilde{a}$ is the predicted TSDF value, 
$a$ is the ground-truth TSDF value, 
$\textbf{p}$ is a point within a set of the ground-truth point clouds $\mathcal{P}$, 
$\tilde{\textbf{p}}$ is a point within a set of predicted point clouds $\tilde{\mathcal{P}}$, 
Prec is the precision metric (\ie, $\text{Prec}\text{=}\text{mean}_{\tilde{\textbf{p}} {\in} \tilde{\mathcal{P}}} (\text{min}_{\textbf{p} {\in} \mathcal{P}} \left | \textbf{p} \text{-} \tilde{\textbf{p}} \right |_{1} {<} 0.05)$), 
and Recall is the recall metric (\ie, $\text{Recall}\text{=}\text{mean}_{\textbf{p} {\in} \mathcal{P}} (\text{min}_{\tilde{\textbf{p}} {\in} \tilde{\mathcal{P}}} \left | \textbf{p} \text{-} \tilde{\textbf{p}} \right |_{1} {<} 0.05)$)

\section{Implementation and Training Scheme}
In this section, we provide clarifications and precision about the implementation and training scheme of our volume fusion network. The size of the n-th initial feature volume $\mathcal{V}_\text{n}^\mathcal{I}$ is $\mathbb{R}^{96(3\text{C}) {\times} 48(\text{D}) {\times} 120(\text{H}) {\times} 160(\text{W})}$. Note that the size of the volumes in the first stage is consistent during the training and testing session, however, we set different resolutions of the feature volume~$\mathcal{V}_\text{n}^{F}$, the pose-invariant scene volume~$\mathcal{V}_\text{n}^{R}$, the unified scene volume$\mathcal{V}^{U}$, and the TSDF scene volume~$\tilde{\mathcal{V}}^\text{TSDF}$. 
The sizes of both the feature volume $\mathcal{V}_\text{n}^{F}$ and the pose-invariant scene volume $\mathcal{V}_{n}^{R}$ are $\mathbb{R}^{32(C) {\times} 160(\text{V}_\text{x}) {\times} 160(\text{V}_\text{y}) {\times} 48(\text{V}_\text{z})}$ for training and $\mathbb{R}^{32(\text{C}) {\times} 640(\text{V}_\text{x}) {\times} 640(\text{V}_\text{y}) {\times} 128(\text{V}_\text{z})}$ for testing and evaluation.

As depicted in~\Fref{fig:fig_arch} of the manuscript, the unified scene volume $\mathcal{V}^{U}$ obeys $\mathbb{R}^{33(\text{C}\text{+}1) {\times} 160(\text{V}_\text{x}) {\times} 160(\text{V}_\text{y}) {\times} 48(\text{V}_\text{z})}$ for a training session and $\mathbb{R}^{33(\text{C}\text{+}1) {\times} 640(\text{V}_\text{x}) {\times} 640(\text{V}_\text{y}) {\times} 128(\text{V}_\text{z})}$ for a testing period. Note that the unified scene volume has one more channel than the pose-invariant scene volume $\mathcal{V}_\text{n}^{R}$. This is because we embed back-projected point clouds from masked depth maps $\hat{\mathcal{Z}}$. After we propagate $\mathcal{V}^{U}$ for depth fusion, the dimensions of the TSDF scene volume $\tilde{\mathcal{V}}^\text{TSDF}$ are $\mathbb{R}^{160(\text{V}_\text{x}) {\times} 160(\text{V}_\text{y}) {\times} 48(\text{V}_\text{z})}$ for a training period and $\mathbb{R}^{640(\text{V}_\text{x}) {\times} 640(\text{V}_\text{y}) {\times} 128(\text{V}_\text{z})}$ for a testing session. 
Note that we set the identical resolution of the final TSDF scene volume as proposed in Murez~\etal~\cite{atlas} for a fair comparison.
The size of the batch per GPU is set as $1$. For training, each batch consists of 45 multi-view images (\ie, $15$ reference views and $30$ neighbor views). We use all images for test. 

\section{Discrete Kernel Rotation}
\label{supp-sec:Discrete Kernel Rotation}

This section describes the details of \emph{PosedConv}. The idea of the \emph{PosedConv} is to rotate the reservoir kernel by using the known camera poses to extract rotation-invariant features (\Sref{subsec:Posed Convolution Layer} of the manuscript).

\begin{figure*}[!t]
\centering
\includegraphics[width=1.0\linewidth]{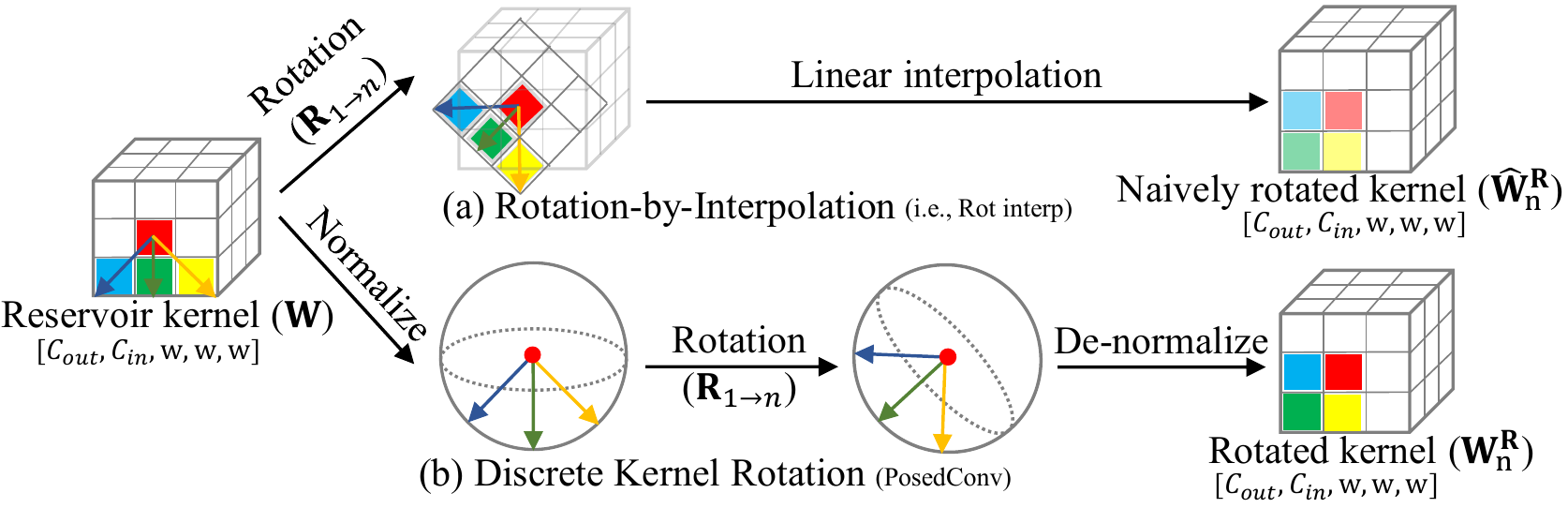}
\caption{\textbf{Discrete Kernel Rotation in our \emph{PosedConv}.} From the reservoir kernel $\mathbf{W}$, (a) Rotation-by-Interpolation produces naively posed kernel $\hat{\mathbf{W}}_\text{n}^\mathbf{R}$, (b) discrete kernel rotation (ours) produces the rotated kernel~$\mathbf{W}_\text{n}^\mathbf{R}$.
}
\label{fig:fig_rotation}
\vspace{+4mm}
\end{figure*}

The rotated kernel can be simply computed by rotation followed by linear interpolation, called Rotation-by-Interpolation, as shown in~\Fref{fig:fig_rotation}-(a). 
However, the naively rotated kernels $\hat{\mathbf{W}}^R_{n}$ often fail to interpolate properly rotated reservoir kernel values because of the different radius from the center to the boundary.
Thus, we design a discrete kernel rotation performed on a unit sphere to minimize the loss of boundary kernel information as shown in~\Fref{fig:fig_rotation}-(b), called \emph{Discrete Kernel Rotation}.

In detail, we first transform the 3D cube into a unit sphere by normalizing its distance from the center of the kernel. Within this unit sphere, the rotated sphere also becomes a unit sphere so that we can alleviate the information drops during the rotation. After rotating the unit sphere, we denormalize the radius of the sphere to form the 3D cube.

This process rotates the reservoir kernel $\mathbf{W}$ of the $\text{n}$-th camera view by using the rotation matrix $\mathbf{R}_{1 \rightarrow \text{n}}$, as depicted in~\Fref{fig:fig_rotation} of the main paper. Concretely, we iteratively apply the discrete kernel rotation for each $\text{n}$-th feature volume $\mathcal{V}^{\mathcal{F}}_\text{n}$, as in~\Fref{fig:fig_arch} of the main paper. Given the reservoir kernel $\mathbf{W}$ and the rotation matrix $\mathbf{R}_{1 \rightarrow \text{n}}$, the algorithm infers the rotated kernel $\mathbf{W}_\text{n}^{\mathbf{R}}$ as in~\Aref{code:Discrete Kernel Rotation}. To densely fill the rotated kernel $\mathbf{W}_\text{n}^{\mathbf{R}}$ with the reservoir kernel $\mathbf{W}$, we need to repetitively apply the reverse warping process.

\begin{algorithm}[t]
\footnotesize
\caption{Discrete Kernel Rotation}
\label{code:Discrete Kernel Rotation}
\begin{algorithmic}[1]
    \Require Reservoir kernel $\mathbf{W} \in \mathbb{R}^{C_\text{out} \times C_\text{in} \times w \times w \times w}$, rotation matrix $\mathbf{R}_{\text{n} \rightarrow 1}$.
	\Procedure{Discrete Kernel Rotation}{$\mathbf{W}$, $\mathbf{R}_{\text{n} \rightarrow 1}$}
    
    \State Declare rotated kernel $\mathbf{W}_\text{n}^{\mathbf{R}} \in \mathbb{R}^{C_\text{out} \times C_\text{in} \times w \times w \times w}$
    	
	\For{$[i, j, k]$ in $\mathbf{W}_\text{n}^{\mathbf{R}}$}
    
	    \State $[X_\text{s}, Y_\text{s}, Z_\text{s}]$ $\leftarrow$ $\text{NORM}([i, j, k]$, $w)$
	    \Comment{$NORM(\cdot)$ in~\Aref{code:Norm}}    
	    
	    
	    \State $[\hat{X}_\text{s}, \hat{Y}_\text{s}, \hat{Z}_\text{s}, 1]^\intercal$ $\leftarrow$ $\mathbf{R}_{\text{n} \rightarrow 1} [X_\text{s}, Y_\text{s}, Z_\text{s}, 1]^\intercal$
	    
	    \State $[\hat{i}, \hat{j}, \hat{k}]$ $\leftarrow$ $\text{DeNorm}([i, j, k], w, [\hat{X}_\text{s}, \hat{Y}_\text{s}, \hat{Z}_\text{s}])$
	    \Comment{DENORM$(\cdot)$ in~\Aref{code:DeNorm}}
	    
        \State $\mathbf{W}_\text{n}^{\mathbf{R}}(i, j, k)$ $\leftarrow$ Bilinear($\mathbf{W}$, $[\hat{i}, \hat{j}, \hat{k}]$)
	    
    \EndFor
	\EndProcedure
\end{algorithmic}
\end{algorithm}
\begin{algorithm}[t]
\footnotesize
\caption{Normalization Function for Discrete Kernel Rotation}
\label{code:Norm}
\begin{algorithmic}[1]
    \Require Voxel coordinate $[i, j, k]$, kernel size $w$.
    \Ensure Normalized image coordinate $[X_\text{s}, Y_\text{s}, Z_\text{s}]$.
	
	\Procedure{Norm}{$[i, j, k]$, w}
	
	\State $r$ $\leftarrow$ $\frac{w \text{-} 1}{2}$ 
	\Comment Kernel radius $r$
	
	\State $\ell$ $\leftarrow$ $\sqrt{((i\text{-}r)^2 + (j\text{-}r)^2 + (k\text{-}r)^2)}$
	
	\State $[X_\text{s}, Y_\text{s}, Z_\text{s}]$ $\leftarrow$ $[\frac{i\text{-}r}{\ell}, \frac{j\text{-}r}{\ell}, \frac{k\text{-}r}{\ell}]$ where $0 \leq X_\text{s}, Y_\text{s}, Z_\text{s} \leq 1.0$
	
	\Return $[X_{s}, Y_{s}, Z_{s}]$
	
	\EndProcedure
\end{algorithmic}
\end{algorithm}
\begin{algorithm}[t]
\footnotesize
\caption{DeNormalization Function for Discrete Kernel Rotation}
\label{code:DeNorm}
\begin{algorithmic}[1]
    \Require Voxel coordinate $[i, j, k] \in \mathbf{W}_\text{n}^{\mathbf{R}}$, kernel size $w$, normalized image coordinate $[\hat{X}_\text{s}, \hat{Y}_\text{s}, \hat{Z}_\text{s}]$ where $0.0 \leq \hat{X}_\text{s}, \hat{Y}_\text{s}, \hat{Z}_\text{s} \leq 1.0$.
    \Ensure Rotated voxel coordinate $[\hat{i}, \hat{j}, \hat{k}] \in \mathbf{W}$.
	
	\Procedure{DeNorm}{$[i, j, k]$, $w$, $[\hat{X}_\text{s}, \hat{Y}_\text{s}, \hat{Z}_\text{s}]$}
	
	\State $r$ $\leftarrow$ $\frac{w \text{-} 1}{2}$ 
	\Comment Kernel radius $r$
	
	\State $\ell$ $\leftarrow$ $\sqrt{(i\text{-}r)^2 \text{+} (j\text{-}r)^2 \text{+} (k\text{-}r)^2}$
	
	\State $[\hat{i}, \hat{j}, \hat{k}]$ $\leftarrow$ $[\ell {\cdot} \hat{X}_\text{s}\text{+}r,\text{ \ } \ell {\cdot}  \hat{Y}_\text{s}\text{+}r, \text{ \ } \ell {\cdot} \hat{Z}_\text{s}\text{+}r]$ where $0 \leq i, j, k \leq w$
	
	\Return $[\hat{i}, \hat{j}, \hat{k}]$
	
	\EndProcedure
\end{algorithmic}
\end{algorithm}

Each voxel $\mathbf{v}'$ at the voxel coordinate $[i, j, k]$, is transformed into its position on the unit sphere $[X_\text{s}, Y_\text{s}, Z_\text{s}, 1]^\intercal$ through the NORM$(\cdot)$ function (Line~\textcolor{blue}{4} in~\Aref{code:Discrete Kernel Rotation}).
The details of the NORM$(\cdot)$ function is described in~\Aref{code:Norm} and this is the fundamental difference between our Discrete Kernel Rotation (\Fref{fig:fig_rotation}-(\textcolor{blue}{b}) of the main paper and \Aref{code:Discrete Kernel Rotation}) and rotation-by-interpolation (\Fref{fig:fig_rotation}-(\textcolor{blue}{a}) of the main paper and \Aref{code:Rotation-by-Interpolation}).
Since this is a reverse warping process, we use the given rotation matrix $\mathbf{R}_{\text{n} \rightarrow 1}$ (Line~\textcolor{blue}{5} in~\Aref{code:Discrete Kernel Rotation}) that is the inverse of the rotation matrix $\mathbf{R}_{1 \rightarrow \text{n}}$ as depicted in~\Fref{fig:fig_rotation} of the main paper.
To do so, we obtain the rotated location $[\hat{X}_\text{s}, \hat{Y}_\text{s}, \hat{Z}_\text{s}, 1]^\intercal$ lying on the unit sphere (Line~\textcolor{blue}{5} in~\Aref{code:Discrete Kernel Rotation}).

\begin{figure*}[!t]
\centering
\includegraphics[width=1.0\linewidth]{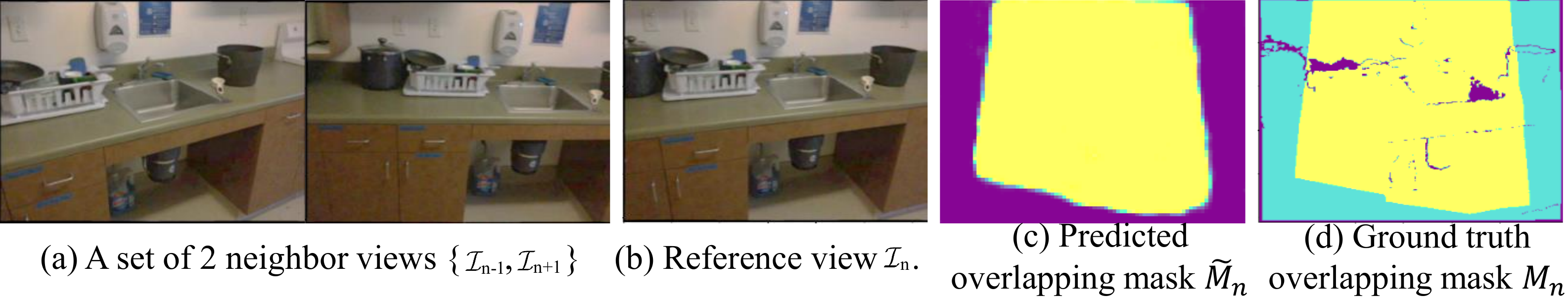}
\caption{\textbf{Illustration of an overlapping mask.} 
Given (a) $2$ neighbor views and (b) a reference frame, 
our multi-view stereo network (\Fref{fig:fig_arch} of the manuscript) infers a depth map $\tilde{\mathcal{Z}}$ and (c)~a overlapping mask $\tilde{\mathcal{M}}$. 
A true overlapping mask ${\mathcal{M}}$ is a binary mask that $1$ represents valid overlapping pixels (yellow pixels in (d)) and $0$ means non-overlapping pixels (cyan pixels in (d)).
We can determine the overlapping pixels as pixels transformed from neighbor images~(a) to the referential image~(b) using true ground truth depth maps $\mathcal{Z}$ and ground truth camera poses $[\mathbf{R}|\mathbf{t}]$.
Since there are unknown true depth values within the ground truth depth maps $\mathcal{Z}$, we cannot determine the overlapping or non-overlapping pixels that are colorized as purple in~(d). 
After training our network, the obtained overlapping mask $\tilde{\mathcal{M}_\text{n}}$ is visualized in~(c) where yellow pixels indicate predicted overlapping pixels and purple pixels mean estimated non-overlapping pixels.
}
\label{fig:fig_conf}
\end{figure*}

To find the corresponding location in the reservoir kernel, we denormalize the rotated location $[\hat{X}_\text{s}, \hat{Y}_\text{s}, \hat{Z}_\text{s}, 1]^\intercal$ and obtain the rotated voxel coordinate $[\hat{i}, \hat{j}, \hat{k}]$ at the reservoir kernel $\mathbf{W}$ (Line~\textcolor{blue}{6} in~\Aref{code:Discrete Kernel Rotation}) where DeNorm$(\cdot)$ function is precisely described in~\Aref{code:DeNorm}.
Since the rotated voxel coordinate is not located outside of the boundary of the reservoir kernel $\mathbf{W}$, we directly apply bilinear interpolation to extract the kernel response at $[\hat{i}, \hat{j}, \hat{k}]$ at the reservoir kernel $\mathbf{W}$ (Line~\textcolor{blue}{7} in~\Aref{code:Discrete Kernel Rotation}).
Contrary to our discrete kernel rotation, naive rotation-by-interpolation must check whether the rotated voxel coordinate $[\hat{i}, \hat{j}, \hat{k}]$ is out of the boundary of the reservoir kernel $\mathbf{W}$, or it sometimes has difficulty in interpolating the reservoir kernel $\mathbf{W}$ into the rotated kernel $\mathbf{W}_\text{n}^{\mathbf{R}}$ (Lines~\textcolor{blue}{8}-\textcolor{blue}{10} in~\Aref{code:Rotation-by-Interpolation}).

Finally, we compute the rotated kernel $\mathbf{W}_\text{n}^{\mathbf{R}}$ for the $\text{n}$-th camera view. As described in~\Sref{subsec:Posed Convolution Layer} of the main paper, we identically apply the conventional 3D convolution operation that is equipped with our rotated kernels $\mathbf{W}_\text{n}^{\mathbf{R}}$.

\begin{table}[!t]
\centering
\resizebox{1.0\linewidth}{!}{
\begin{tabular}{lcccc}
\Xhline{2\arrayrulewidth}
\multirow{2}{*}{Method} & \multicolumn{4}{c}{Evaluation} \\ \cline{2-5}
& AbsRel & RMSE & $\mathcal{L}_{1}$ & F-score \\ \Xhline{2\arrayrulewidth}

3D Conv & .058 & .231 & .166 & .460 \\

Rot interp & .056 & .223 & .159 & .487 \\ 

Disc K Rot & \textbf{.049} & \textbf{.164} & \textbf{.141} & \textbf{.508} \\ \Xhline{2\arrayrulewidth}

\end{tabular}
}
\vspace{2mm}
\caption{\textbf{Ablation study of PosedConv.} We compare original convolution layer~(\ie, 3D Conv), and the Rotation-by-Interpolation~(\ie, Rot interp), and our PosedConv~(\ie, Disc K Rot). It shows that the kernel rotation is much effective thank the original 3D Conv, but our \emph{PosedConv} further improve the quality of the 3D scene reconstruction.
}
\label{table:ablation-posedconv-supp}
\vspace{+4mm}
\end{table}
\begin{algorithm}[!t]
\footnotesize
\caption{Rotation-by-Interpolation}
\label{code:Rotation-by-Interpolation}
\begin{algorithmic}[1]
    \Require Reservoir kernel $\mathbf{W} \in \mathbb{R}^{C_\text{out} \times C_\text{in} \times w \times w \times w}$, rotation matrix $\mathbf{R}_{n \rightarrow 1}$.
    
	\Procedure{Rotation-by-Interpolation}{$\mathbf{W}$, $\mathbf{R}_{\text{n} \rightarrow 1}$}
    
    \State Declare rotated kernel $\mathbf{W}_\text{n}^{\mathbf{R}} {\in} \mathbb{R}^{\text{C}_\text{out} \times \text{C}_\text{in} \times w \times w \times w}$
    	
	\For{$[i, j, k]$ in $\mathbf{W}_\text{n}^{\mathbf{R}}$}
	
	    \State $r$ $\leftarrow$ $\frac{w \text{-} 1}{2}$ 
	    \Comment Kernel radius $r$
	    
	    \State $[X, Y, Z]$ $\leftarrow$ $[i \text{-} r, j \text{-} r, k \text{-} r]$
	    \Comment Signed distance

	    \State $[\hat{X}, \hat{Y}, \hat{Z}, 1]^\intercal$ $\leftarrow$ $\mathbf{R}_{\text{n} \rightarrow 1} [X, Y, Z, 1]^\intercal$
	    
	    \State $[\hat{i}, \hat{j}, \hat{k}]$ $\leftarrow$ $[\hat{X}\text{+}r, \hat{Y}\text{+}r, \hat{Z}\text{+}r])$
	    
	    \If{$[\hat{i}, \hat{j}, \hat{k}] \notin \mathbf{W}$}
	    \Comment{Boundary check}
	        \State $[\hat{i}, \hat{j}, \hat{k}]$ $\leftarrow$ $[\widetilde{i}, \widetilde{j}, \widetilde{k}]$ 
	        \State where $[\widetilde{i}, \widetilde{j}, \widetilde{k}] {\in} \mathbf{W}$ and $[\hat{i}, \hat{j}, \hat{k}]$ is nearest to $[\widetilde{i}, \widetilde{j}, \widetilde{k}]$.
        \EndIf
	    
	    \State $\mathbf{W}_\text{n}^{\mathbf{R}}(i, j, k)$ $\leftarrow$ Bilinear($\mathbf{W}$, $[\hat{i}, \hat{j}, \hat{k}]$)
    \EndFor
	\EndProcedure
\end{algorithmic}
\end{algorithm}

To validate the necessity of our discrete kernel rotation, we conduct an ablation study as in~\Tref{table:ablation-posedconv-supp}. Alongside the results in~\Tref{table:ablation-posedconv} of the manuscript, we additionally report the accuracy when we use naive Rotation-by-Interpolation (\ie, Rot interp). It shows that the rotating the kernel in both ways (Rot interp and discrete kernel rotation) improves the quality of depth and 3D geometry, but our PosedConv further achieves the higher accuracy than that of Rotation-by-Interpolation.

\section{Overlapping Mask}
In this section, we present an example figure of an overlapping mask as in~\Fref{fig:fig_conf}. In the first stage of our network, multi-view stereo, we utilize three neighbor views to infer a depth map and an overlapping mask. This overlapping mask is used to filter out the uncertain depth values at the specific pixels that have no corresponding pixels in the neighbor views. As shown in~\Fref{fig:fig_conf}, our network properly infers the overlapping mask in a referential camera viewpoint. Note that the referential image is in between two adjacent views, the overlapping region is usually located at the center of the referential images.

\section{Combined Ablation Results.}
To clearly show influences from our contributions, we merge ablation results in the manuscripts as in~\Tref{table:merged-ablation}.
The quality of the reconstruction is largely improved with both \textit{PosedConv} and overlapping masks, simultaneously. 
This is because overlapping masks are designed to filter out the uncertain depth values that can hurt the quality of 3D reconstruction done by PosedConv. 
In conclusion, our two novel contributions, PosedConv and overlapping masks, are complementary and lead to precise 3D scene reconstruction.
\begin{table*}[!t]
\centering
\begin{tabular}{cccccccccc}

\Xhline{2\arrayrulewidth}

\multicolumn{5}{c}{Preserve (\cmark)} & \multicolumn{5}{c}{Performance} \\ 

\cline{1-4} \cline{6-10}

Depth & 
\multicolumn{2}{c}{Conv type} & \multirow{2}{*}{Overlap} & &
\multicolumn{2}{c}{2D depth} & &  \multicolumn{2}{c}{3D geometry} \\ 
\cline{2-3} \cline{6-7} \cline{9-10}

fusion & 
Conv & PosedConv & & & 
AbsRel & RMSE & & $\mathcal{L}_\text{1}$ & F-score \\

\Xhline{2\arrayrulewidth}

& 
& & & & 
.061 & .248 & & .162 & .499 \\ 

& 
& \cmark & & & 
.060 & .245 & & .160 & .495  \\ 

\hline


\cmark &
& \cmark & & & 
.060 & .238 & & .162 & .475 \\


\cmark &
\cmark & & \cmark & & 
.058 & .231 & & .166 & .460 \\

\cmark &
& \cmark & \cmark & & 
\textbf{.049} & \textbf{.164} & & \textbf{.141} & \textbf{.508} \\

\Xhline{2\arrayrulewidth}
 
\end{tabular}
\caption{\textbf{Merged ablation results on ScanNet dataset.}}
\label{table:merged-ablation}
\end{table*}

\begin{table*}[!t]
\centering
\begin{tabular}{lcccccc}
\Xhline{2\arrayrulewidth}

\multirow{2}{*}{Method} & \multicolumn{2}{c}{2D Depth} & & \multicolumn{2}{c}{3D Geometry} \\ \cline{2-3} \cline{5-6}

 & AbsRel & RMSE & & Acc & F-score \\

\Xhline{2\arrayrulewidth}


CNMNet (ECCV'20) & .161 & .361 & & .398 & .149 \\ 

NeuralRecon (CVPR'21) & .155 & .347 & & .100 & \textbf{.228} \\


\hline 

VolumeFusion (ours) & \textbf{.140} & \textbf{.320} & & \textbf{.085} & .217 \\ \Xhline{2\arrayrulewidth}

\end{tabular}
\caption{\textbf{Quantitative results on 7-Scenes dataset.}}
\label{table:7scenes-rebuttal}
\end{table*}

\section{Additional Results}
We conduct quantitative evaluations on the 7-Scenes dataset (Shotton~\etal,~CVPR'13) in~\Tref{table:7scenes-rebuttal}. Similarly, our network achieves state-of-the-art performance against the recent approaches. For a fair comparison, we strictly follow the assessment pipelines of these concurrent approaches~(\cite{neural_recon} and Long~\etal in ECCV'20).

\end{document}